%% file: arxiv.tex
\documentclass[letterpaper,12pt]{article} 

\usepackage{subfiles}
\usepackage{amsmath}
\usepackage{amssymb}
\usepackage[authoryear]{natbib}
\usepackage[dvips]{epsfig}
\usepackage{dcolumn}
\usepackage{enumerate}
\usepackage{hhline}
\usepackage{dsfont}
\usepackage{afterpage}
\usepackage{arydshln}
\usepackage{graphicx}
\usepackage{color}
\usepackage[usenames,dvipsnames]{xcolor}
\usepackage{rotating}
\usepackage[]{hyperref} %breaklinks
\usepackage{xr}
\usepackage[percent]{overpic}
\usepackage{subfig}
\usepackage[]{caption}

\usepackage{diagbox}
\usepackage{graphicx}
\usepackage{wrapfig}
\usepackage{lscape}
\input epsf
\usepackage{fontenc}
\usepackage{setspace}
\usepackage{bm}
\usepackage{slashbox}
\usepackage{lscape}
\usepackage{multirow}
\usepackage{eurosym}
\usepackage{titlesec}
\RequirePackage[linesnumbered,ruled]{algorithm2e} % algorithms
\RequirePackage{algpseudocode}
\usepackage{mathtools} 
\epsfverbosetrue

\input{defs.tex}

\usepackage[verbose=true,letterpaper]{geometry}
\titlespacing*\section{0pt}{0pt plus 4pt minus 2pt}{0pt plus 2pt minus 2pt}
\titlespacing*\subsection{0pt}{0pt plus 4pt minus 2pt}{0pt plus 2pt minus 2pt}
\titlespacing*\subsubsection{0pt}{0pt plus 4pt minus 2pt}{0pt plus 2pt minus 2pt}

\usepackage{paralist}

\renewenvironment{itemize}[1]{\begin{compactitem}#1}{\end{compactitem}}

\makeatletter
\def\@seccntformat#1{\@ifundefined{#1@cntformat}%
	{\csname the#1\endcsname\quad}  % default
	{\csname #1@cntformat\endcsname}% enable individual control
}
\let\oldappendix\appendix %% save current definition of \appendix
\renewcommand\appendix{%
	\oldappendix
	\newcommand{\section@cntformat}{\appendixname~\thesection\quad}
}

\usepackage[verbose=true,letterpaper]{geometry}
\AtBeginDocument{
  \newgeometry{
    textheight=9in,
    textwidth=6in,
    top=1in,
    headheight=12pt,
    headsep=25pt,
    footskip=30pt
  }
}

\renewenvironment{abstract}%
{%
  \vskip 0.075in%
  \centerline%
  {\large\bf Abstract}%
  \vspace{0.5ex}%
  \begin{quote}%
}
{
  \par%
  \end{quote}%
  \vskip 1ex%
}

\renewcommand{\normalsize}{%
  \@setfontsize\normalsize\@xpt\@xipt
  \abovedisplayskip      7\p@ \@plus 2\p@ \@minus 5\p@
  \abovedisplayshortskip \z@ \@plus 3\p@
  \belowdisplayskip      \abovedisplayskip
  \belowdisplayshortskip 4\p@ \@plus 3\p@ \@minus 3\p@
}

\normalsize
\renewcommand{\small}{%
  \@setfontsize\small\@ixpt\@xpt
  \abovedisplayskip      6\p@ \@plus 1.5\p@ \@minus 4\p@
  \abovedisplayshortskip \z@  \@plus 2\p@
  \belowdisplayskip      \abovedisplayskip
  \belowdisplayshortskip 3\p@ \@plus 2\p@   \@minus 2\p@
}
\renewcommand{\footnotesize}{\@setfontsize\footnotesize\@ixpt\@xpt}
\renewcommand{\scriptsize}{\@setfontsize\scriptsize\@viipt\@viiipt}
\renewcommand{\tiny}{\@setfontsize\tiny\@vipt\@viipt}
\renewcommand{\large}{\@setfontsize\large\@xiipt{14}}
\renewcommand{\Large}{\@setfontsize\Large\@xivpt{16}}
\renewcommand{\LARGE}{\@setfontsize\LARGE\@xviipt{20}}
\renewcommand{\huge}{\@setfontsize\huge\@xxpt{23}}
\renewcommand{\Huge}{\@setfontsize\Huge\@xxvpt{28}}

\providecommand{\section}{}
\renewcommand{\section}{%
  \@startsection{section}{1}{\z@}%
                {-2.0ex \@plus -0.5ex \@minus -0.2ex}%
                { 1.5ex \@plus  0.3ex \@minus  0.2ex}%
                {\large\bf\raggedright}%
}
\providecommand{\subsection}{}
\renewcommand{\subsection}{%
  \@startsection{subsection}{2}{\z@}%
                {-1.8ex \@plus -0.5ex \@minus -0.2ex}%
                { 0.8ex \@plus  0.2ex}%
                {\normalsize\bf\raggedright}%
}
\providecommand{\subsubsection}{}
\renewcommand{\subsubsection}{%
  \@startsection{subsubsection}{3}{\z@}%
                {-1.5ex \@plus -0.5ex \@minus -0.2ex}%
                { 0.5ex \@plus  0.2ex}%
                {\normalsize\bf\raggedright}%
}
\providecommand{\paragraph}{}
\renewcommand{\paragraph}{%
  \@startsection{paragraph}{4}{\z@}%
                {1.5ex \@plus 0.5ex \@minus 0.2ex}%
                {-1em}%
                {\normalsize\bf}%
}
\providecommand{\subparagraph}{}
\renewcommand{\subparagraph}{%
  \@startsection{subparagraph}{5}{\z@}%
                {1.5ex \@plus 0.5ex \@minus 0.2ex}%
                {-1em}%
                {\normalsize\bf}%
}

\providecommand{\@maketitle}{}
\renewcommand{\@maketitle}{%
  \vbox{%
    \hsize\textwidth
    \linewidth\hsize
    \vskip 0.1in
    \centering
    {\LARGE\bf \@title\par}
      \def\And{%
        \end{tabular}\hfil\linebreak[0]\hfil%
        \begin{tabular}[t]{c}\bf\rule{\z@}{24\p@}\ignorespaces%
      }
      \def\AND{%
        \end{tabular}\hfil\linebreak[4]\hfil%
        \begin{tabular}[t]{c}\bf\rule{\z@}{24\p@}\ignorespaces%
      }
      \begin{tabular}[t]{c}\bf\rule{\z@}{24\p@}\@author\end{tabular}%
    \vskip 0.3in \@minus 0.1in
  }
}

\makeatother

\usepackage{titlesec}

\usepackage{scalerel,stackengine}
\stackMath
\newcommand\reallywidehat[1]{%
\savestack{\tmpbox}{\stretchto{%
  \scaleto{%
    \scalerel*[\widthof{\ensuremath{#1}}]{\kern-.6pt\bigwedge\kern-.6pt}%
    {\rule[-\textheight/2]{1ex}{\textheight}}%WIDTH-LIMITED BIG WEDGE
  }{\textheight}% 
}{0.5ex}}%
\stackon[1pt]{#1}{\tmpbox}%
}

\begin{document}
\setlength{\abovedisplayskip}{0.15cm}
\setlength{\belowdisplayskip}{0.15cm}
%\pagestyle{empty}
%\singlespacing
\subfile{title}
\setcounter{equation}{0}
\renewcommand{\theequation}{\arabic{equation}}
\subfile{content}
\end{document}

%% file: defs.tex
\def \dsE {\text{$\mathds{E}$}}
\def \dsR {\text{$\mathds{R}$}}

\def \cdnn {\text{$c_{\mbox{\scriptsize{DNN}}}$}}
\DeclareMathOperator{\diag}{diag}

    \def \mB {\text{\boldmath$B$}}
    
\def \dvec {\text{\boldmath$d$}}

    \def \mI {\text{\boldmath$I$}}

    \def \mP {\text{\boldmath$P$}}
    
    \def \mR {\text{\boldmath$R$}}
    \def \mS {\text{\boldmath$S$}}
    
\def \uvec {\text{\boldmath$u$}}

\def \xvec {\text{\boldmath$x$}}    
\def \yvec {\text{\boldmath$y$}}    \def \mY {\text{\boldmath$Y$}}
\def \zvec {\text{\boldmath$z$}}    \def \mZ {\text{\boldmath$Z$}}

\def \betavec         {\text{\boldmath$\beta$}}

\def \deltavec        {\text{\boldmath$\delta$}}

\def \varepsilonvec   {\text{\boldmath$\varepsilon$}}
\def \zetavec         {\text{\boldmath$\zeta$}}

\def \thetavec        {\text{\boldmath$\theta$}}
\def \varthetavec     {\text{\boldmath$\vartheta$}}

\def \lambdavec       {\text{\boldmath$\lambda$}}
\def \muvec           {\text{\boldmath$\mu$}}

\def \xivec           {\text{\boldmath$\xi$}}

\def \rhovec          {\text{\boldmath$\rho$}}

\def \psivec          {\text{\boldmath$\psi$}}

\def \mUpsilon {\mathbf{\Upsilon}}

%% file: title.tex
%\begin{titlepage} %del for first page

\title{Marginally Calibrated Response Distributions for End-to-End Learning in Autonomous Driving}

\author{Clara Hoffmann and Nadja Klein$^\ast$}%\footnote{Nadja Klein is the Emmy Noether Research Group Leader in Statistics and Data Science  at Humboldt-Universit\"at zu Berlin and Clara Hoffmann in her research team. Correspondence should be directed to~Prof.~Dr.~Nadja Klein at Humboldt Universit\"at zu Berlin, School of Business and Economics, 
%Unter den Linden 6, 10099 Berlin. Email: nadja.klein@hu-berlin.de.\\
%\textbf{Acknowledgments:} Support by the Deutsche Forschungsgemeinschaft (DFG, German Research Foundation) through the Emmy Noether grant KL 3037/1-1 is gratefully acknowledged. The authors would like to thank Cornelius Hoffmann for providing storage and server infrastructure for the data and computations.}}
%$^1$Humboldt  Universit\"at zu Berlin\\
%$^2$National University of Singapore\\
%$^3$Melbourne Business School, University of Melbourne
\date{}
\maketitle
\noindent

\begin{abstract}
End-to-end learners for autonomous driving are deep neural networks that predict the instantaneous steering angle directly from images of the ahead-lying street. These learners must provide reliable uncertainty estimates for their predictions in order to meet safety requirements and initiate a switch to manual control in areas of high uncertainty. Yet end-to-end learners typically only deliver point predictions, since distributional predictions are associated with large increases in training time or additional computational resources during prediction. To address this shortcoming we investigate efficient and scalable approximate inference for the implicit copula neural linear model of \citet{KleNotSmi2020} in order to quantify uncertainty for the predictions of end-to-end learners. The result are densities for the steering angle that are marginally calibrated, i.e.~the average of the estimated densities equals the empirical distribution of steering angles. To ensure the scalability to large $n$ regimes, we develop efficient estimation based on variational inference as a fast alternative to computationally intensive, exact inference via Hamiltonian Monte Carlo. 
We demonstrate the accuracy and speed of the variational approach in comparison to Hamiltonian Monte Carlo on two end-to-end learners trained for highway driving using the comma2k19 data set. The implicit copula neural linear model delivers accurate calibration, high-quality prediction intervals and allows to identify overconfident learners. Our approach also contributes to the explainability of black-box end-to-end learners, since predictive densities can be used to understand which steering actions the end-to-end learner sees as valid. 
\end{abstract}
\vspace{10pt}
 
\noindent
{\bf Keywords}: {Autonomous driving}; 
{calibration}; 
{deep neural network}; 
{distributional regression}; 
{implicit copula}; 
{neural linear models}; 
{probabilistic forecasting}; 
{uncertainty quantification}; 
{variational inference}. 

\vspace{10pt}
\noindent $^\ast$ {Nadja Klein is the Emmy Noether Research Group Leader in Statistics and Data Science at Humboldt-Universit\"at zu Berlin and Clara Hoffmann is part of her research team. Correspondence should be directed to~Prof.~Dr.~Nadja Klein at Humboldt Universit\"at zu Berlin, School of Business and Economics, 
Unter den Linden 6, 10099 Berlin. Email: nadja.klein@hu-berlin.de.\\
\vspace{10pt} \\
\textbf{Acknowledgments:} Support by the Deutsche Forschungsgemeinschaft (DFG, German Research Foundation) through the Emmy Noether grant KL 3037/1-1 is gratefully acknowledged. The authors would like to thank Cornelius Hoffmann for providing storage and server infrastructure for the data and computations.}
\vspace{10pt} \\

%% file: content.tex
\section{Introduction}
In recent years there have been immense advancements in autonomous driving, but progressing from mere driver's assistance to fully autonomous drivers still poses great safety challenges. Due to the high costs associated with wrong decisions of autonomous drivers, models have to be extremely accurate and safe in a wide range of driving scenarios. As a response to safety and scalability issues of traditional autonomous driving systems, a new family of models, called end-to-end learners, has emerged in the last decade. End-to-end learners predict steering angles directly from images or videos of the ahead-lying street of a driving car using a single deep neural network (DNN). These models usually only provide a point prediction for the steering angle. This makes it impossible to quantify uncertainty or address potential overconfidence of a learner. 

Obtaining accurate predictive uncertainty measures for the steering angles is essential to reliably assess the safety of an end-to-end learner and can also be leveraged to initiate a switch to human control in areas of high uncertainty to avoid crashes. Recently, there has been much progress in obtaining reliable uncertainty estimates for a DNN's prediction. But the vast training set sizes as well as temporal and computational limitations during prediction in autonomous driving pose challenges to these methods. As a solution we propose to obtain predictive densities for the steering angle building on the marginally calibrated deep distributional regression model of \cite{KleNotSmi2020}. We label this approach the implicit copula neural linear model (IC-NLM), to highlight its connection to the class of neural linear models (NLMs). NLMs comprise DNNs, where the last layer is augmented to a Bayesian linear regression model. The IC-NLM is based on the implicit copula \citep[Sec.~5 of][]{Nel2006} of a vector
of transformed response variables that arises from a NLM. The resulting copula allows to model highly flexible relations between the feature vector and the response densities. The copula is combined with a non-parametrically estimated marginal distribution for the
observed response variable to ensure certain calibration properties.  \\
In the IC-NLM, predictive densities are obtained by computing the complete posterior distributions of the last layer weights. For this, we develop stable, exact estimation via Hamiltonian Monte Carlo \citep[HMC;][]{Nea2011} and approximate estimation via variational inference (VI). Here, HMC is used as a stable alternative to the MCMC sampler of \cite{KleNotSmi2020} based on Metropolis-Hastings and Gibbs steps and we use it to verify that VI delivers accurate results. While HMC delivers exact posteriors, it is only feasible for moderately sized data sets and as the sample size $n$ grows, convergence issues arise and run-times become long. Therefore, HMC is not a scalable option for realistic autonomous driving models with large $n$. As a solution to this issue we introduce VI as our methodological innovation. This ensures the scalability of the IC-NLM to huge data sets as it is the case in end-to-end learning and also for the full comma2k19 data set \citep{SchEdeHad2018} with $n > 300{,}000$ observations on which we base our analysis. 

We train two end-to-end learners for highway driving on the comma2k19 data and quantify the uncertainty of the steering angle predictions using the IC-NLM and a number of state-of-the-art benchmark methods.
The comparison with HMC illustrates that the proposed VI approach is accurate both for small and large sample sizes. In a benchmark study, we show that the IC-NLM produces reliable, informative predictive densities and exhibits better calibration than two competing methods commonly used for measuring predictive uncertainty in this context. 

The remainder of the paper is structured as follows. We start out with the basic idea and development of end-to-end learners as well as the unique challenges encountered when quantifying uncertainty for their predictions in Section~\ref{sec:background}. We also introduce the data for training our end-to-end learners in this section. In Section~\ref{sec:NLM} we present the IC-NLM, starting with the definition of NLMs and continuing with reviewing the construction of the implicit copula version of the NLM. Section~\ref{sec:inference} develops both exact and approximate inference through HMC and VI to estimate the IC-NLM.  Section~\ref{sec:results} contains our detailed analysis to quantify uncertainty in end-to-end learners. The final Section \ref{sec:discussion} concludes.

\section{Background and challenges in end-to-end learning}\label{sec:background}
Traditionally, autonomous driving approaches consist of modularized models that rely on object-classification and object-tracking paired with if-else behavior rules \citep{CheSefKorXia2015}. But these models still struggle to navigate vehicles autonomously without human intervention.  One reason for the failures of traditional models are limited training sets. Providing sufficient training data involves large costs: object-detection models require bounding boxes for each object in a single training image. Providing bounding boxes for the training data can only be done manually by humans, making large training sets very costly to produce. Small training sets in turn lead to driving models that are not robust against the diversity of situations encountered on real-world roads. 

\begin{figure}[htbp]
    \centering
    \includegraphics[width= 11cm]{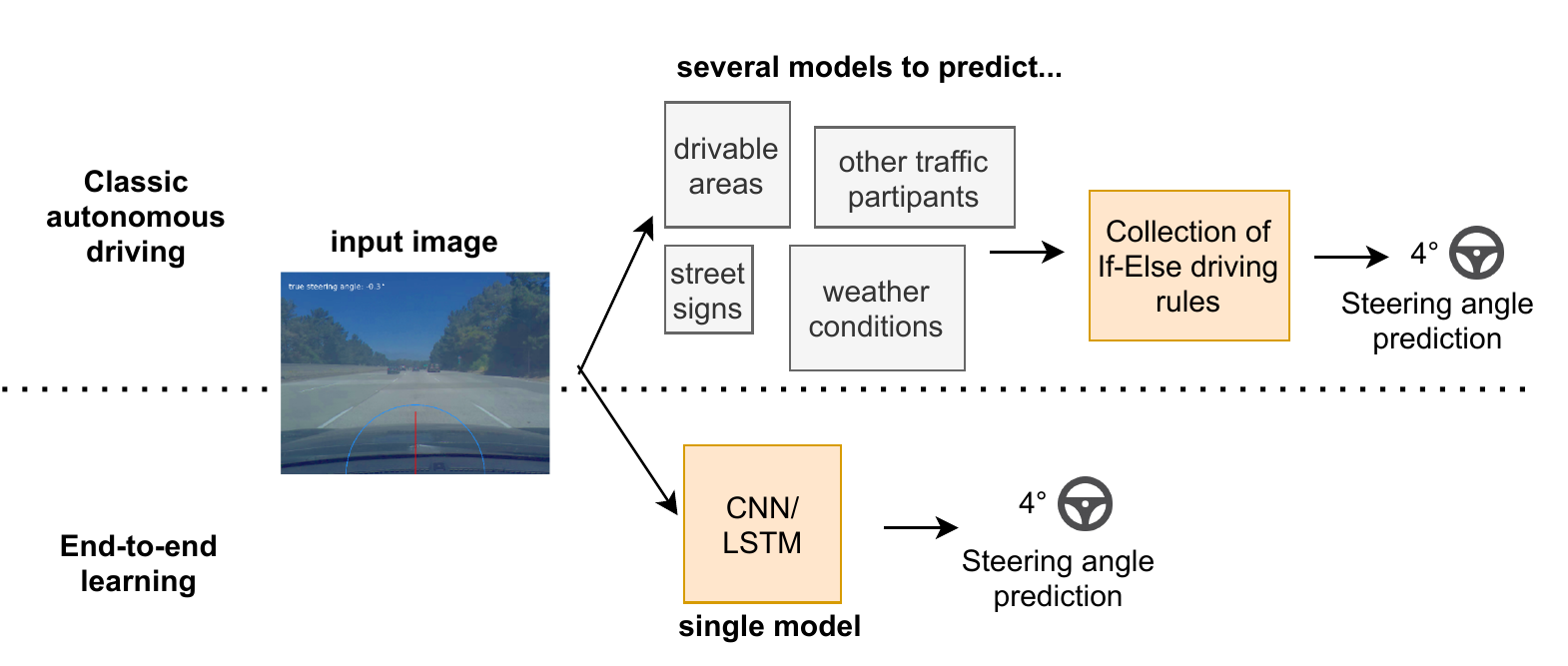}
    \caption{Classic autonomous driving models vs.~end-to-end learning}
    \label{fig:etelearning}
\end{figure}

Eliminating manual annotation opens new prospects for ultra-large training sets and motivates the evolution to so-called end-to-end learners. These models directly translate sensory inputs to driving instructions through a single, non-modularized model. Typically, the steering angle is predicted from images of the ahead-lying street through a single DNN. The first end-to-end learners were convolutional neural networks (CNN)s that predict steering angles directly from the images of a forward-facing camera on a driving car \citep{BojTesDwo2016}. Nowadays, end-to-end learners have evolved to also take spatial and time dimensions into account \citep{XuGaoYuTre2016, ChiMu2017, AmiSolKarRus2018} and use e.g.~long short-term memory (LSTM) networks.
The workflow under the classic paradigm in autonomous driving and under end-to-end learning is illustrated in Figure~\ref{fig:etelearning}.

Formally, an end-to-end learner is a mapping from images of the ahead-lying road to the steering angle. Let $\boldsymbol{Y} = (Y_1,..., Y_n)^\top$ to be a random vector of $n$ steering angles, from which we observe realizations $\boldsymbol{y} = (y_1,...y_n)^\top$. The corresponding observed features $\boldsymbol{x}_i$,  $\boldsymbol{x} = ( \boldsymbol{x}_1,...,\boldsymbol{x}_n)^T$ are tensors that contain the image pixels. Typically all pixels are stored in three-dimensional tensors. The dimensions correspond to position of the pixels along the width, height and color channels of the image. An end-to-end learner is a mapping $f(\boldsymbol{x}_i) = \hat{y}_i$ that can be used to predict steering angle also from new feature values.  

End-to-end learners typically only deliver point predictions $\hat{y}_i=E[Y_i\mid\xvec_i]$ without any uncertainty quantification. However, it is important to ensure the safety of end-to-end learners. Hence, one is interested in predicting not only $\hat y_i$  but rather the entire predictive density  $p(y^\ast\mid\xvec^\ast)$ at an arbitrary steering angle $y^\ast$ given an arbitrary input image $\xvec^\ast$. The latter can be used to identify regions of high uncertainty, initiate a switch to manual control or act as an early alert system for wrong predictions. Overconfident end-to-end learners can also be identified through $p(y^\ast\mid\xvec^\ast)$ since they will associate wrong predictions with low predictive variance. Due to its attractiveness for autonomous driving applications, obtaining suitable estimators for $p(y^\ast\mid\xvec^\ast)$ has recently become the topic of extensive research. However, most current methods exhibit computational difficulties or suffer from a lack of accuracy. The reason is that many methods for computing $p( y^\ast\mid\xvec^\ast)$ conflict with the unique hardware and prediction speed requirements in autonomous vehicles. Besides, existing methods often do not produce densities that are accurate, reliable and consistent with the data due to unrealistic model assumptions. 

Obtaining predictive densities for end-to-end learners in autonomous driving comes with unique challenges. To ensure safety, predictive densities should be well calibrated. Calibration is an essential criterion to ensure reliability of predictive uncertainty. But currently, few methods for estimating $p(y^\ast\mid\xvec^\ast)$ based on DNNs produce calibrated estimates. We give an overview of important calibration notions in Section~\ref{sec:calibration}. Another challenge is to obtain flexible predictive distributions $p(y^\ast\mid\xvec^\ast)$ for end-to-end learners at low cost. Predictive densities have to be complex (e.g.~multimodal, skewed) to reflect the existence of several valid steering options. Training has to be relatively cheap since large models with huge training sets are needed. Prediction has to be fast to ensure real-time prediction but also hardware-efficient since only limited computational resources are available in an autonomous vehicle \citep{LinZhaHsu2018}. We discuss and investigate how well current methods for obtaining predictive densities in end-to-end learning handle these challenges in our analysis in Section~\ref{sec:literature_overview} and compare them to the IC-NLM.

\subsection{Notions of calibration and reliability of predictive distributions}\label{sec:calibration}
In the machine learning literature, calibration is often understood as the reliability of prediction intervals. For the IC-NLM we will adhere to the more statistical calibration notions introduced by \cite{GneBalRaf2007}, namely \emph{marginal} and \emph{probabilistic} calibration. Reliable prediction intervals will arise as a natural by-product to these calibration types, as we will see later. 

Formally, marginal calibration can be expressed as follows: Assume that we compute predictive densities for the elements of a stochastic process $\{Y_1, Y_2, ... \}$. The corresponding probabilistic predictions are predictive cumulative distribution functions (CDF)s, which are continuous, strictly increasing and collected in a sequence $(\hat{F}_i)_{i \, \epsilon \, \mathbb{N}}$. The true CDFs of the data-generating process are denoted as $(F_i)_{i \, \epsilon \, \mathbb{N}}$. Following \citet{GneBalRaf2007}, marginal calibration of $(\hat{F}_i)_{i \, \epsilon \, \mathbb{N}}$ relative to $(F_i)_{i \, \epsilon \, \mathbb{N}}$ occurs if the asymptotic limits of the average true distribution and the average predictive CDFs exist and equal each other, i.e.
\begin{equation*} \label{eq:marg_cal}
    \lim_{n \rightarrow \infty} \Big( \frac{1}{n}\sum_{i=1}^n F_i(y) \Big) = \lim_{n \rightarrow \infty} \Big( \frac{1}{n}\sum_{i=1}^n \hat{F}_i(y) \Big), \; \; \forall \, y \, \epsilon \,\mathbb{R}.
\end{equation*}
Note that in our case of cross-sectional data, there is no ordering in the sequence.
 However, as noted in 
\cite{gneiting+br07}, the above framework can 
still provide related empirical 
notions of marginal calibration. \cite{gneiting+r13}
develop such a notion and show that marginal calibration can be defined as equality of the marginal distribution of the observation and the expected forecast distribution.  
In the context of end-to-end learning, marginal calibration can be interpreted as a stable distribution over driving trajectories. For example, when training an end-to-end learner for lane-keeping on highways, we would expect that most steering trajectories are straight and no extreme curves are driven, except when leaving the highway. Probabilistic predictions that are not marginally calibrated could place too much probability mass on extreme steering angles, which evidently is inconsistent with the distribution over steering angles that we observe in practice. We will see in Section \ref{sec:results} that the IC-NLM produces marginally calibrated predictive densities for in-sample observations per construction. 

Another notion of calibration is probabilistic calibration.
The predictive densities are probabilistically calibrated if, on average, the probability that we observe $Y_i \leq y$ under the predictive CDF $\hat{F}_i$ converges almost surely to the probability to observe $Y_i \leq y$ under $F_i$, i.e.
\begin{equation*} \label{eq:prob_cal}
    \frac{1}{n} \sum_{i=1}^n F_i \circ \hat{F}_i^{-1}(p) \overset{a.s.}{\to} p, \; \; \forall \, p \, \epsilon \, (0,1),
\end{equation*}
where $\overset{a.s.}{\to}$ denotes almost sure convergence. Even though the IC-NLM does not provide theoretical guarantees for probabilistic calibration, we will see in Section~\ref{sec:results} that its predictions still perform quite well under this aspect. 

Predictive densities can also be used to compute prediction intervals and identify potentially wrong predictions. Following the definition of \citet{PeaZakBri2018}, an $1-\alpha$\% prediction interval is an interval $[\hat{y}_{i,\mbox{\scriptsize{LB}}}, \hat{y}_{i,\mbox{\scriptsize{UB}}}]$ such that an observation $y_i$ falls into this interval with probability of at least $1-\alpha$\%
\begin{equation*}\label{eq:prediction_interval}
   P( \hat{y}_{i,\mbox{\scriptsize{LB}}} \leq y_i \leq \hat{y}_{i,\mbox{\scriptsize{UB}}}) \geq 1 -\alpha.
\end{equation*}
The coverage rates of the prediction intervals should be accurate so that the width of the prediction interval at a given level can be used to quantify uncertainty of the current prediction. 

\subsection{
Probabilistic models and uncertainty quantification }\label{sec:literature_overview}
In current methods for probabilistic predictions with DNNs there exists a trade-off between computational cost and predictions accuracy.
Probabilistic models often possess a substantially higher number of parameters than their non-probabilistic counterparts such that researchers are confronted with long computing times or increased hardware requirements. This is the case for e.g.~Bayesian neural networks (BNN)s, that learn probability distributions over all network weights. Even state-of-the art algorithms for estimating BNNs \citep{BluCorKavWie2015} lead to at least twice as many parameters as in a non-Bayesian DNN. 
In contrast, most non-Bayesian, ensemble-inspired approaches that estimate $p(y^\ast\mid\xvec^\ast)$ require several parallel DNN evaluations. Given that the hardware in autonomous vehicles is restricted, evaluating a sufficient number of DNNs in parallel is not always feasible. Some methods that suffer from these drawbacks are the ensemble approach of \cite{LakPriBlu2017} and MC-Dropout as proposed in \cite{GalGha2016} and applied to autonomous driving by \cite{AmiSolKarRus2018,MicKwiGal2018}. 
In addition, empirically both (pseudo-) ensembles and BNNs do generally not produce calibrated probabilistic predictions. This is often attributed to a lack of model expressivity and diversity \citep{KulFenErm2018,ZhaDalSab2019}.

A few approaches are exempt from the aforementioned trade-off. Uncertainty of end-to-end learners can be quantified by discretizing the steering angle into bins \citep{XuGaoYuTre2016, ChiMu2017}. Thereby the regression problem is turned into a classification problem. The resulting class probabilities can easily be used for uncertainty quantification, but are often not calibrated \citep{GupPleSun2017} and explanations to this problem can be found in \cite{ZhaDalSab2019}.
Beyond that, distributional models that can capture aspects of the response distribution beyond
the mean have evolved also in deep learning. For instance, heteroscedastic Gaussian models \citep{KenGal2017}, deep versions of quantile regression \citep{RodPer2020} or generalized linear models \citep{TraNguNotKoh2020} have been suggested. 
Mixture density networks \citep[MDNs;][]{Bis1994,UriMurLar2013} also produce predictive densities, are easy to train and require little modification from a non-probabilistic DNN. The only drawback is that the resulting predictive densities are empirically often not calibrated.

\subsection{Data}\label{sec:data}

\begin{figure}
\begin{center}
\centering
    \subfloat{{\includegraphics[width= 6cm]{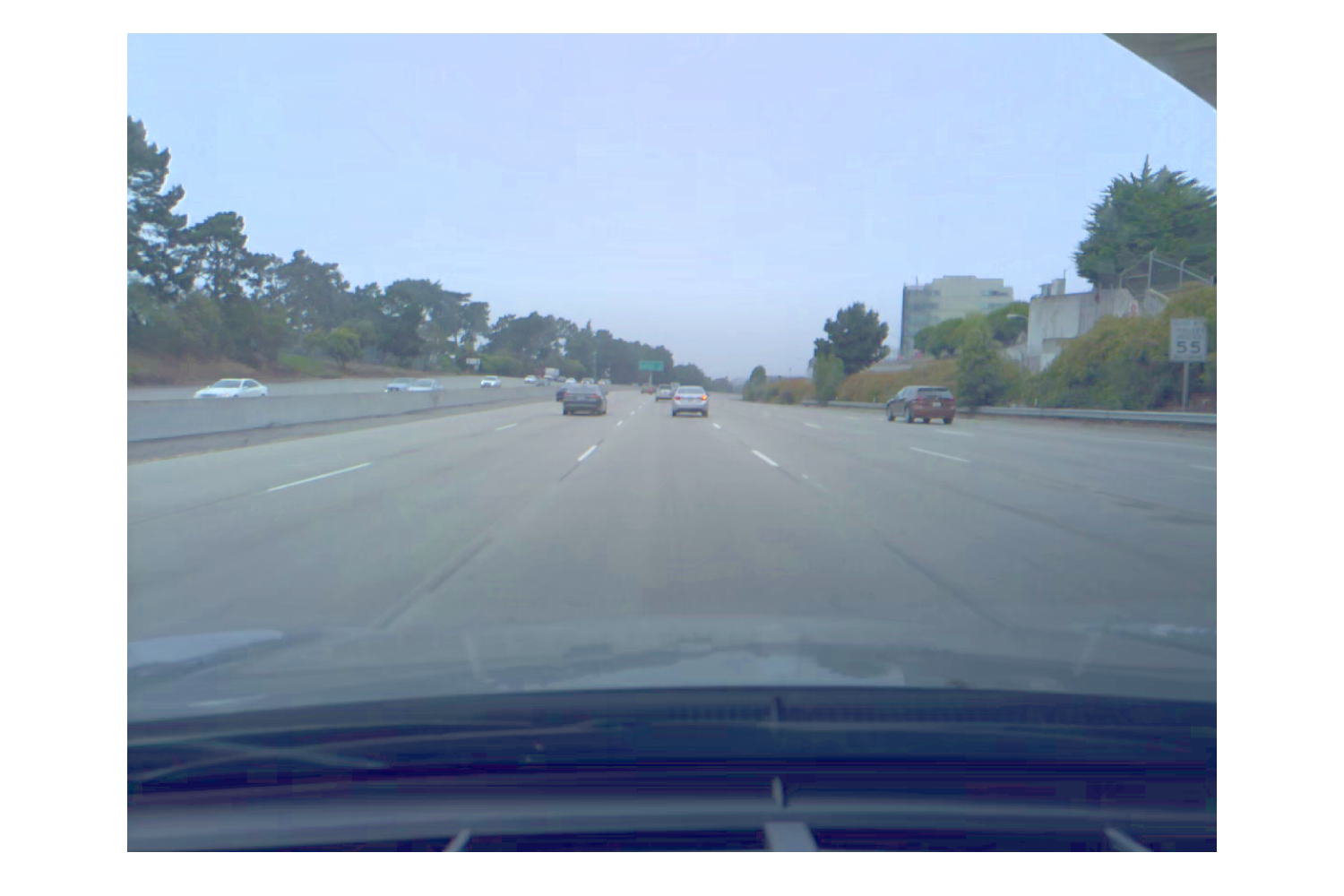} }}
    \vspace{-0.5cm}
    \subfloat{{\includegraphics[width= 6cm]{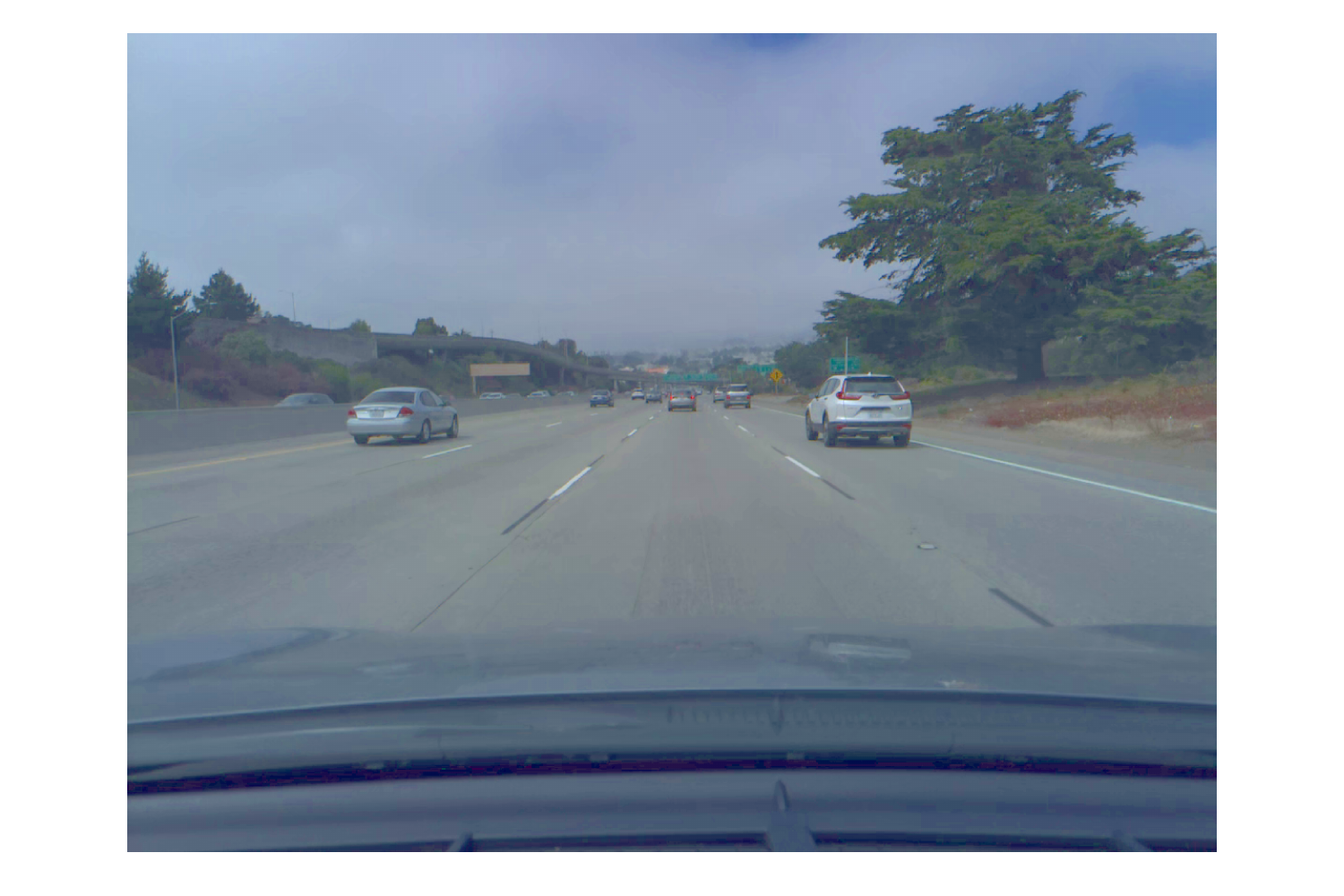} }}
\end{center}
\caption{Example images from the comma2k19 data set}
\label{fig:examples_commaai}
\end{figure}

We demonstrate the IC-NLMs speed and accuracy by quantifying the uncertainty of end-to-end learners trained on the comma2k19 data set \citep{SchEdeHad2018}. The data contains over 33 hours of driving footage on highways in California, collected in 2{,}019 video files. A camera mounted on the windshield records the ahead-lying road and several sensory inputs are measured simultaneously. The comma2k19 data is particularly suitable to train an end-to-end learner, since on highways lane markings are clearly visible, making it easy for a CNN to use them as a steering orientation. Two example frames from the data are depicted in Figure~\ref{fig:examples_commaai}.
We train an end-to-end learner on a small and large version of the data set. The small data set is used to explore how much uncertainty is present in an end-to-end learner for simple highway driving. The large version is used solely to explore the scalability of our VI approach to the IC-NLM. The small version is manually cleaned from erratic driving behavior and lane changes. It can be used for learning lane keeping and is split into 43{,}736 training and 10{,}472 validation observations. An end-to-end learner trained on this data could be directly employed on real highways. We use it to provide a realistic quantification of uncertainty of end-to-end learners for highway driving. This data scenario also ensures that the IC-NLM performs well when the relation between covariates and the response is highly informative and only disturbed by little noise. An end-to-end learner trained on the small data set could not be used for more complex environments such as cities or crowded highways, as this would require significantly more manually cleaned training data. The small data scenario is also not sufficient to explore the computational gains of VI. 

Since large, manually cleaned driving data sets are not easily available, we instead use the full, raw data to explore the scalability of the IC-NLM. This large data set is divided into a train and validation set with 355{,}543 training and 155{,}386 validation observations. This large version will be used to investigate the benefits of using VI instead of HMC. A learner trained on the large data version could probably not be employed directly on a highway, since important information that guides driving decisions such as lane changes is not passed to the model (e.g. route information or rear views). But the size of the large data version is much more realistic for real-world autonomous driving applications\footnote{Additionally, much interest lies in directly using raw data for training end-to-end learners \citep[(compare~e.g.][]{XuGaoYuTre2016} because cleaning data sets from erratic or unwanted driving behavior is extremely time consuming. Although some semi-supervised methods have been developed to overcome this issue (see \cite{AbbHajKar2019} for an overview), hand-picking training examples is still a common step in training end-to-end learners.}. Predicting uncertainty for the large data also illustrates our methodological contribution: Employing VI instead of HMC reduces the computation time by several days when using the full data.

\section{The Implicit Copula Neural Linear Model}\label{sec:NLM}
\subsection{Primer on DNNs}
Typically, end-to-end learners for autonomous driving are DNNs that map from images of the ahead-lying road directly to the steering angle.
Generally, DNNs can be used to approximate arbitrary continuous functions $f^*( \boldsymbol{X}^\ast) = Y^\ast$ mapping a random vector of features $\boldsymbol{X}^\ast$
(not necessarily tabular but e.g.~images or text) to a scalar response $Y^\ast$. DNNs consist of layers of neurons, where each neuron receives the neuron activations from the previous layer as inputs. The neurons of the network jointly implement a complex nonlinear mapping from the input to the output through weight matrices of linear transformations and nonlinear activation functions. This mapping is learned from the data by adapting the weights of each neuron using a technique called error backpropagation \citep{RumHinWil1986}. 

For end-to-end learners, the last layers of a DNN typically consist of fully connected, dense layers. The output $f_\zetavec(\xvec^\ast):=f^{(k)}, \, k \, \epsilon \, \{1,...,K\}$ of the $k$-th layer of such a dense layer for a single observation with feature $\xvec^\ast$ can be represented as 
\begin{align*} 
   f_\zetavec(\xvec^\ast)= f^{(k)} = g^{(k)}(\boldsymbol{Z}^{(k)}(\xvec^\ast)\boldsymbol{W}^{(k)} + \boldsymbol{b}^{(k)}),
\end{align*}
where $\boldsymbol{Z}^{(k)}$ are the outputs of the previous ($k-1$-th) layer, $\boldsymbol{W}^{(k)}$ is a weight matrix,  $\boldsymbol{b}^{(k)}$ is a bias vector, $g^{(k)}$ is a (non-)linear activation function and $\zetavec$ is the set of all weights and biases of the DNN up to the $k$-th hidden layer. The weights and biases of all layers are determined by minimizing a empirical loss criterion, such as the mean squared error loss (MSE), $\sum_{i=1}^n(f_\zetavec(\xvec_i)-y_i)^2$, based on  a training set $\mathcal{D} = \{(\boldsymbol{x}_i, y_i)\}_{i= 1}^n$ of features and responses  
 \citep[we refer interested readers to][for a further discussion on how to determine the weights of a DNN and more information on DNNs, including regularization]{GooBencou2016,PolSok2017}. In a regression setting, minimization based on the MSE is equivalent to to assuming a homoscedastic Gaussian
model (thus assuming that the responses are indeed conditionally Gaussian), although minimization does not require parametric model assumptions. 

\subsection{Neural linear models}
The class of NLMs comprises Bayesian linear models, where the features are deep basis functions learned by a DNN with an identity activation function in the last layer \citep{SnoRipSwe2015, RiqTucSno2018,ObeRas2019,PinGorNal2019}. Assume we have trained a DNN on a training set $\mathcal{D}$. We denote the $n\times p$ matrix of outputs from the last hidden layer as  $\boldsymbol{B}_{{\zetavec}}(\boldsymbol{x}) = \lbrack\psivec_\zetavec(\boldsymbol{x}_1)\mid\ldots| \psivec_\zetavec(\boldsymbol{x}_n)\rbrack^T \in \mathbb{R}^{n \times p}$, where
$\psivec_\zetavec(\cdot)$ is the
vector of $p$ basis functions defined by the last hidden layer. Then, the NLM is of the form
\begin{equation}\label{eq:NLM}
    \boldsymbol{y} = \boldsymbol{B}_{{\zetavec}}(\boldsymbol{x})\boldsymbol{\beta}+\beta_0 + \boldsymbol{\varepsilon},
\end{equation}
where $\boldsymbol{\beta}=(\beta_1,\ldots,\beta_p)^\top\in\mathds{R}^p$ is the vector of regression coefficients (or weights), $\boldsymbol{\varepsilon}=(\varepsilon_1,\ldots\varepsilon_n)^\top$ is the vector of i.i.d.~error terms, $\boldsymbol{\varepsilon} \sim N(\boldsymbol{0}, \sigma^2 \boldsymbol{I})$ and $\beta_0 \in \mathbb{R}$ is the intercept (also called bias in the machine learning literature). Both, the vector of weights  $\boldsymbol{\beta}$ and the bias $\beta_0$ are equipped with a prior in an NLM.  This allows to perform Bayesian inference over the models parameters and compute the posterior of $\boldsymbol{\beta}$ and subsequently the predictive densities $p(y^\ast\mid\xvec^\ast)$ via the posterior predictive densities (see Section \ref{subsec:preddens}). 

Recently, it has become popular to obtain Bayesian inference over the coefficient vector by expressing the likelihood of $\boldsymbol{y}$ in closed form and learning all weights of the DNN by maximizing the likelihood for the response given the NLM specification. This is done by formulating an appropriate loss function based on the likelihood that typically contains hyperparameters \citep{SnoRipSwe2015, RiqTucSno2018, PinGorNal2019}. However, we do not make use of this approach in the IC-NLM but separate the learning process from Bayesian inference. Most of the existing NLMs use simple, conjugate priors, which in turn results in insufficiently complex, unimodal predictive densities $p(y^\ast\mid\xvec^\ast))$. These densities are not appropriate for modeling complex uncertainty scenarios in autonomous driving. The IC-NLM overcomes this issue by employing an implicit copula. The latter allows the whole distribution of the steering angle to vary flexibly with the features. 
Construction of the  IC-NLM in the next section will be based on a NLM and we will in this context also give details on prior specifications and posterior inference.

\subsection{Derivation of the Implicit Copula Neural Linear Model}\label{sec:IC-NLM}

To ensure marginally calibrated densities, \citet{KleSmi2019} and~\cite{SmiKle2020} introduce a new approach to distributional
regression that uses a copula decomposition constructed from the so-called inversion method~\citep{Nel2006}. \citet{KleNotSmi2020} outline how to extend their approach to deep learning regression which we refer to as IC-NLM. We review the key ideas of this method and the adaptations for our application in the following.

\subsubsection{Calibrated copula process}
Copula models with regression margins for multivariate responses $\mY\in\dsR^D$, $D>1$ have been widely used in the literature \citep[see e.g.][]{PitChaKoh2006,SonLiYua2009,CraSab2012,KleKne2016b}.
However,
another usage of a copula with regression data is to capture the dependence between multiple
observations on a single dependent variable $Y$, conditional on the feature values; which in our case are the steering angles and corresponding feature image tensors. Doing so, defines a \emph{copula process} \citep{WilGha2010} on the feature space, which \citet{SmiKle2020} call a \emph{regression copula} because of its dependence on the features. When combined with a flexible marginal distribution
for $Y^\ast$, the authors show that it specifies a tractable and scalable distributional (i.e.~probabilistic) regression model where the entire distribution of $Y^\ast$ varies with the features. \citet{KleNotSmi2020} use this idea when the regression copula is the implicit copula  of the
joint distribution of a \emph{pseudo} response,   and these follow a  DNN regression model. As a consequence, the pseudo responses can be obtained through density transformations from the observed response $\mY$, as detailed in Section~\ref{sec:deepcopula} below.  We now first explain the copula construction and its relation to other statistical probabilistic models.

Sklar's theorem~\citep{Skl1959} states that the $n$-dimensional distribution of $\mY\mid\xvec$ can be written as
\begin{equation*}
    p(\yvec\mid \xvec) = c^\dagger(F_{Y_1}(y_1\mid \xvec_1),\ldots,F_{Y_n}(y_n\mid \xvec_n)\mid\xvec)\prod_{i=1}^n p_{Y_i}(y_i\mid x_i),
\end{equation*}
where the $n$-dimensional copula with  density $c^\dagger$ is a copula process on the covariate space, $\yvec$
and $F_{Y_i}(y_i\mid\xvec_i)$ is the marginal distribution function of $Y_i\mid\xvec_i$ with density $p_{Y_i}$. Both, the copula and the marginal distributions are typically unknown, and it is common to make the copula dependent on some copula parameters $\thetavec$. We follow \citet{KleNotSmi2020} and use $\cdnn(\uvec\mid\xvec,\thetavec)$ with $u_i = F_{Y_i}(y_i)$ (cf.~Section~\ref{sec:deepcopula}). One further tractable but effective
simplification is to allow the covariates to only affect the dependent variable through the
copula function~\citep{KleSmi2019} and to calibrate the distribution of $Y_i\mid x_i$ to its invariant margin, so
that  $F_{Y_i}\equiv F_Y$ has density $p_{Y_i}(y_i\mid x_i)=p_Y(y_i)$. This margin can be
estimated non-parametrically.  Thus, in the IC-NLM the response has the joint density
\begin{equation}\label{eq:jdensy}
    p(\yvec\mid\xvec,\thetavec) = \cdnn(u_1,\ldots,u_n\mid\xvec,\thetavec)\prod_{i=1}^n p_Y(y_i).
\end{equation}
The marginal invariance assumption may seem counter-intuitive in a first place given the usual conditional formulation of a regression model for $Y_i\mid\xvec_i$ through a univariate marginal model. Still,  it is valid approach for which indeed  $\mY$ is dependent on $\xvec$ in the joint
distribution defining a flexible distributional regression model, see \citet{SmiKle2020} for details.

\subsubsection{Copula construction}\label{sec:deepcopula}

The basis for the construction of the IC-NLM of \citet{KleNotSmi2020} is a DNN trained to predict the density-transformed pseudo responses $\mZ$, with an identity activation function in the last layer. Let $\tilde Z_i$ be the $i$-th pseudo response, $i=1,\ldots,n$. Then $\tilde Z_i$ can be modeled using the output layer of the DNN plus some Gaussian noise $\varepsilon_i\sim N(0,\sigma^2)$, i.e.~$\tilde Z_i=f_i^{(K)}+\varepsilon_i$, where $f_i^{(K)}$ denotes the output when passing the $i$-th feature observation to the DNN. Then, from \eqref{eq:NLM}, the vector $\tilde\mZ = (\tilde Z_1,\ldots,\tilde Z_n)^\top$ follows the linear model \begin{equation}\label{eq:NLM2}
\tilde\mZ=\mB_\zetavec(\xvec)\betavec+\varepsilonvec,
\end{equation}
where the intercept $\beta_0$ was excluded since it will not be identified in the copula. The weights $\zetavec$ can be easily obtained by training the DNN to predict the pseudo response (see Section~\ref{sec:inference} for details on how to fix them). Thus, \eqref{eq:NLM2} is a linear model with design matrix $\mB_\zetavec$ and  vector of weights $\betavec$. Efficient estimates for $\betavec$ are produced via regularization with the conditionally Gaussian prior as a shrinkage prior, i.e.
\[
\betavec\mid\thetavec,\sigma^2\sim N(\bm{0},\sigma^2 \mP(\thetavec)^{-1})\,.
\]
The (sparse) precision matrix $\mP(\thetavec)$ is a function of the copula parameters $\thetavec$.
The density $\cdnn$ is then derived by integrating $\betavec$ out in \eqref{eq:NLM2} \citep[see][for a detailed derivation]{KleSmi2019}. This results in a Gaussian copula \citep{Son2000} with density
  \begin{equation}
 c_{\mbox{\tiny DNN}}(\uvec\mid\xvec,\thetavec) = 
 \frac{p(\zvec\mid\xvec,\sigma^2,\thetavec)}{\prod_{i=1}^n p(z_i\mid\xvec,\sigma^2,\thetavec)}=
 \frac{\phi_n(\zvec;\bm{0},\mR(\xvec,\thetavec))}{\prod_{i=1}^n \phi_1(z_i)}\,,
 \label{eq:copdens}
 \end{equation}
 where 
 \begin{equation}
 \mR(\xvec,\thetavec) = \mS(\xvec,\thetavec)\left( \mI+\mB_\zetavec(\xvec)\mP(\thetavec)^{-1}\mB_\zetavec(\xvec)^\top \right) \mS(\xvec,\thetavec)\,,
 \label{eq:omega}
  \end{equation}
 $z_i=\Phi_1^{-1}(u_i)$, $\zvec=(z_1,\ldots,z_n)^\top$, and $\phi_n(\cdot;\bm{0},\mR)$ and $\phi_1$ are the densities of $N_n(\bm{0},\mR)$
  and $N(0,1)$ distributions, respectively. The random variables $Z_i$ are standardized versions of $\tilde Z_i$, $\bm{Z}=(Z_1,\ldots,Z_n)^\top=\sigma^{-1}\mS(\xvec,\thetavec)\tilde{\bm{Z}}$, where 
$\mS(\xvec,\thetavec)=\mbox{diag}(s_1,\ldots,s_n)$ is a diagonal scaling matrix
with elements $s_i=(1+ \psivec_{\zetavec}(\xvec_i)^\top \mP(\thetavec)^{-1}
 \psivec_{\zetavec}(\xvec_i))^{-1/2}$, which ensures that
  $Z_i\mid\xvec,\sigma^2,\thetavec \sim N(0,1)$.

\subsubsection{Shrinkage for deep regression copulas}

\begin{figure}[ht] 
    \centering
    \hspace*{1.5cm}\includegraphics[scale = 0.45]{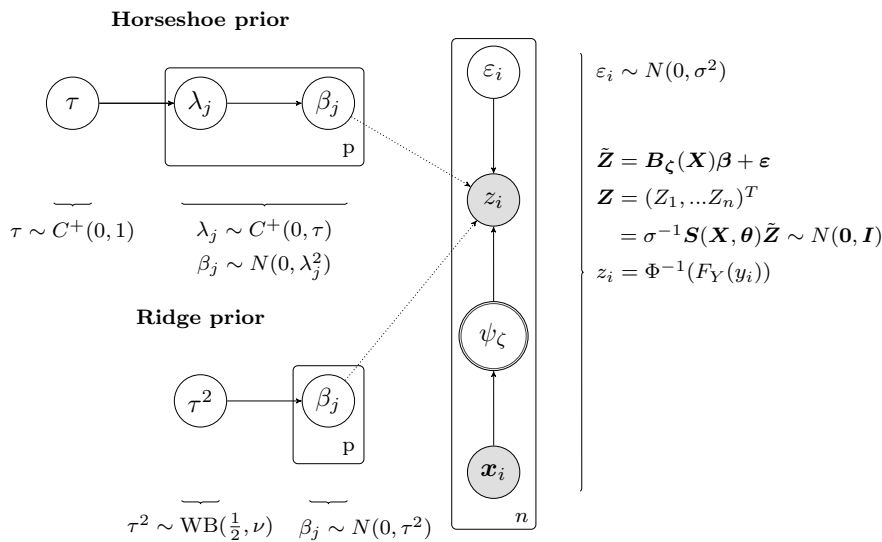}
    \vspace{-2.3cm}
    \caption{The IC-NLM as a graphical model. Gray arrows indicate that only one of the dependencies can hold at a time, i.e. either the ridge or horseshoe prior.}
    \label{fig:graphical_model}
\end{figure}

 For specific choices of $\mP(\thetavec)$, we compare the two shrinkage priors employed in \citet{KleNotSmi2020}, namely the ridge and horseshoe prior. 
 \paragraph*{Ridge} The ridge prior is one of the simplest forms of shrinkage priors, where $\beta_j\mid\tau^2\sim N(0,\tau^2)$, $j=1,\ldots,p$ and a hyperprior is used on the variance $\tau^2$.  For this we employ the robust and principled choice of scale-dependent priors of~\cite{KleKne2016a}, which corresponds to a Weibull  prior, $\tau^2\sim\mbox{WB}(1/2,\nu)$ with scale parameter $\nu$. Predictive densities were rather robust with respect to the actual value of $\nu$ and we follow \citet{KleNotSmi2020} and set $\nu=2.5$ in our analysis.
\paragraph*{Horseshoe} The horseshoe prior is attractive due to its robustness, local adaptivity and
analytical properties~\citep{CarPol2010}.
It is a scale mixture of the hierarchical form $\beta_j\mid\lambda_j\sim N(0,\lambda_j^2)$, 
with $\pi_0(\lambda_j\mid\tau)=\mbox{Half-Cauchy}(0,\tau^2)$ and 
$\pi_0(\tau)=\mbox{Half-Cauchy}(0,1)$. Compared to the ridge prior, this prior is a global-local shrinkage prior, which allows for weight-specific local shrinkage for each coefficient in addition to overall regularization through $\tau^2$.  
Even though the dimension of $\betavec$ is quite small in our application ($p = 10$, see Section \ref{sec:results}), regularizing the weights can make the posterior response densities more robust to noise and enhance predictive quality as we show in Section~\ref{sec:results}. While the ridge prior introduces only one further scalar parameter $\thetavec=\lbrace \tau^2\rbrace$, the horseshoe prior comes with $p+1$ hyperparameters $\thetavec=\lbrace \lambda_1,\ldots,\lambda_p,\tau^2\rbrace$ and the corresponding correlation matrices for both priors are
\begin{equation*}\begin{aligned}
\mR(\xvec,\bm{\theta})&=\mS(\bm{x},\bm{\theta})\left(\mI+\tau^2 \mB_\zetavec(\xvec)\mB_\zetavec(\xvec)^\top\right)
\mS(\bm{x},\bm{\theta})
\\
\mR(\xvec,\bm{\theta})&=\mS(\bm{x},\bm{\theta})\left(\mI+\mB_\zetavec(\xvec)\diag(\lambdavec)^2 \mB_\zetavec(\xvec)^\top\right)
\mS(\bm{x},\bm{\theta}).
\end{aligned}\end{equation*}
Allowing for more sophisticated shrinkage of each regression coefficient introduces additional parameters to the model, which in turn can slow down estimation. However, our results show that allowing for local shrinkage in addition to global regularization significantly enhances predictive performance and the model's ability to identify potentially high-error predictions. This is why we argue that employing a more complex shrinkage prior through the horseshoe is worth the additional computational cost.
The full hierarchical IC-NLM is illustrated as a graphical model in Figure \ref{fig:graphical_model}.

\subsection{Estimation of the IC-NLM}
Algorithm \ref{alg:fullinference} summarizes how estimation for the IC-NLM is realized. 

\begin{algorithm}[H] \label{alg:fullinference}
%\setstretch{1}
 \SetKwInOut{Input}{Input}
 \SetKwInOut{Output}{Output}
 \Input{Features $\boldsymbol{x}$, responses $\boldsymbol{y}$ for training and new features $\boldsymbol{x}_0$ for prediction.}
 \Output{Posterior densities for the response $\hat p({y}_0\mid\boldsymbol{x}_0)$}
 Estimate the empirical density $\hat{F}_Y$ from the observed responses $\boldsymbol{y}$ and transform $\boldsymbol{y}$ to $\boldsymbol{z}$ via $z_i = \Phi_1^{-1}(\hat{F}_Y(y_i))$. \\
 Train a DNN to predict the transformed responses $\boldsymbol{z}$ from $\boldsymbol{x}$ and save the weights up the last hidden layer as $\boldsymbol{\zeta}$. \\
 Predict the outputs $\mB_{\boldsymbol{\zeta}}(\boldsymbol{x})$ of the last hidden layers of the DNN. \\
 Estimate $p(\boldsymbol{\beta}, \boldsymbol{\theta} \mid \boldsymbol{x}, \boldsymbol{y})$ by MCMC or VI as described in Section \ref{sec:inference}. \\
 Compute posterior predictive densities $\hat p({y_0}\mid \boldsymbol{x}_0)$ as described in Section \ref{subsec:preddens}. 
 \caption{Estimation of the IC-NLM} \label{alg:icnlm}
 \end{algorithm}

Step 1 involves estimating the marginal distribution $\hat F_Y$ via a kernel density estimator (KDE). Later, we will use a non-parametric KDE with a Gaussian kernel \citep{Rac2008}, see Section \ref{sec:results}.
 Steps~2 and 3 are dependent upon the choice of architecture, which we discuss later
in Section \ref{sec:results}.  Step~4  requires
evaluation of the likelihood, which is given by the copula decomposition at~\eqref{eq:jdensy}. To do so directly requires evaluation 
of the copula
density at~\eqref{eq:copdens}, which is computationally infeasible 
in general because of the need to invert
the $n\times n$ matrix $\mR$.
\cite{KleSmi2019} solve this problem by instead using
the likelihood conditional also on $\betavec$, which is
\begin{equation}\begin{aligned}\label{eq:clike}
p(\yvec\mid\xvec,\betavec,\thetavec)&= p(\zvec\mid\xvec,\betavec,\thetavec)\prod_{i=1}^n 
\frac{p_{Y}(y_i)}{\phi_1(z_i)}\\
&=\phi_n\left(\zvec;S(\xvec,\thetavec)B_\zetavec(\xvec)\betavec,
S(\xvec,\thetavec)^2\right)\prod_{i=1}^n\frac{p_{Y}(y_i)}{\phi_1(z_i)}\,,
\end{aligned}\end{equation}
and can be evaluated in $O(n)$ operations because $S(\xvec,\thetavec)$ is diagonal. We present exact and approximate Bayesian inference for this step in Section \ref{sec:inference}. Computation of Step 5 is based on \eqref{eq:clike} and described next.
All steps are implemented in Python and the code is available on \href{https://github.com/clarahoffmann/commaai}{github} .

\subsubsection{Predictive densities and beyond}\label{subsec:preddens}
The desired predictive uncertainty for the steering angle at new feature values $\boldsymbol{x}_0$ will be based on the posterior predictive density $p(y_0 \mid \boldsymbol{x}_0, \boldsymbol{x}, \boldsymbol{y})$ of the IC-NLM, which is for a new value $y_0$ is given by
\begin{equation}\label{eq:preddens}
    p(y_0 \mid \boldsymbol{x}_0, \boldsymbol{x}, \boldsymbol{y}) = \int p(y_0 \mid \boldsymbol{x}_0,\xvec,\boldsymbol{\beta}, \boldsymbol{\theta}) p(\boldsymbol{\beta}, \boldsymbol{\theta}\mid \boldsymbol{x}, \boldsymbol{y})d(\boldsymbol{\beta}, \boldsymbol{\theta}),
\end{equation}
where we use $p(y_0 \mid \boldsymbol{x}_0, \boldsymbol{x}, \boldsymbol{\beta}, \boldsymbol{\theta})$ instead of $p(y_0 \mid \boldsymbol{x}_0, \boldsymbol{x}, \boldsymbol{\theta})$ to avoid direct computation of $\mR(\xvec, \thetavec)$. 
If $z_0=\Phi_1^{-1}(F_Y(y_0))$, then $\left| \frac{dz_0}{dy_0} \right|=\frac{p_Y(y_0)}{\phi_1(z_0)}$, and by 
changing variables from $y_0$ to $z_0$, estimates $\hat p$ of \eqref{eq:preddens} can be obtained    via
\begin{equation*}\label{eq:predictivedensest}
    \hat{p}(y_0 \mid \boldsymbol{x}_0) =\frac{p_Y(y_0)}{\phi_1(z_0)}\hat{p}(z_0\mid\xvec_0,\xvec,\yvec)= \frac{p_Y(y_0)}{\phi_1(\Phi_1^{-1}(F_Y(y_0))} \frac{1}{\hat{s_0}} \phi_1 \Big( \frac{\Phi_1^{-1}(F_Y(y_0)) -  \hat{m}(\boldsymbol{x}_0)}{\hat{s_0}} \Big).
\end{equation*}
Here, $\hat{m}(\xvec_0)= \hat{s}_0\psi_{{\zeta}}(\boldsymbol{x}_0)^T\hat{\boldsymbol{\beta}}$ and $\hat s_0=(1+ \psi_{\zetavec}(\xvec_0)^\top \mP(\hat\thetavec)^{-1}
 \psivec_{\zetavec}(\xvec_0))^{-1/2}$ based on the posterior mean VI estimates $\hat\betavec,\hat\thetavec$ in case of approximate posterior estimation (see Section \ref{sec:VB}) or from the MCMC output $\{(\betavec^{(1)}, \thetavec^{(1)}), ..., (\betavec^{(J)}, \thetavec^{(J)}) \}$  via  $\hat{m}(\xvec_0)=\psivec_{{\zeta}}(\boldsymbol{x}_0)^T( \frac{1}{J}\sum_{j=1}^J s_0^{[j]} \boldsymbol{\beta}^{[j]})$ and $\hat s_0 = \frac{1}{J}\sum_{j=1}^J s_0^{[j]}$  in case of exact posterior estimation (see Section \ref{sec:MCMC}).  Note that $\hat{m}(\xvec_0)$ is an estimator for $\dsE[Z_0\mid\xvec_0,\xvec,\yvec]$.

Later in our analysis, we will also make use of the posterior predictive variance, ${\mbox{Var}}(Y_0\mid\xvec_0,\xvec,\yvec) = \dsE[Y_0^2\mid\xvec_0,\xvec,\yvec] - \dsE[Y_0\mid\xvec_0,\xvec,\yvec]^2$, with  posterior predictive mean 
\begin{equation*}\begin{aligned}
\dsE[Y_0\mid \xvec_0,\xvec,\yvec]&=\int\dsE[Y_0\mid \xvec_0,\xvec,\betavec,\thetavec]p(\betavec,\thetavec\mid\xvec,\yvec){d}(\betavec,\thetavec)\\
&=\int F_Y^{-1}(\Phi_1(z_0))\frac{1}{s_0}\phi_1\left(\frac{z_0-s_0 \psivec_{\zetavec}(\xvec_0)^\top\betavec}{s_0}\right) {d}z_0\,,
\end{aligned}\end{equation*}
and
\begin{equation*}\begin{aligned}
\dsE[Y_0^2\mid \xvec_0,\xvec,\yvec]&=\int \left(F_Y^{-1}(\Phi_1(z_0))\right)^2\frac{1}{s_0}\phi_1\left(\frac{z_0-s_0 \psivec_{\zetavec}(\xvec_0)^\top\betavec}{s_0}\right) {d}z_0\,.
\end{aligned}\end{equation*} 
Both integrals involve univariate numerical integration only. Estimators for the integrands are obtained in the same fashion as for the predictive densities. If the true steering angle is available, the prediction error can be measured by the squared or absolute deviation to $\hat{\dsE}(Y_0\mid\xvec_0,\xvec,\yvec)$. Prediction intervals can also easily be obtained from the predictive densities. Usually symmetric prediction intervals are used where the lower bound is the response value at which the predictive CDF is smaller than $\alpha/2$ and the upper bound is the response value at which the CDF is larger than $1-\alpha/2$.
%The next section introduces the MCMC and VI schemes in detail and motivate the usefulness of VI.

\section{Exact and approximate posterior estimation}\label{sec:inference} 
In this section, we present a stable MCMC scheme for posterior estimation of the IC-NLM which is applicable for moderate data sets. It is based on HMC, since the MCMC sampler of \citet{KleNotSmi2020} based on Gibbs and Metropolis-Hastings updates was too sticky to produce reliable results in our application. Next, we develop a VI approach as an approximate alternative for posterior estimation with large-scale data sets and highly parameterized models. Even though MCMC delivers  estimates in both of our data scenarios,  it is plagued by slow convergence and long runtimes as $n$ grows large. This is an issue for autonomous driving applications, where sample sizes can become extremely large. The VI approach is much faster than MCMC and sufficiently accurate, as we show in Section \ref{sec:results}. In the following we denote $\varthetavec = \{ \betavec, \thetavec\}$ the set of all model parameters and set $p_{\varthetavec}=\dim(\varthetavec)$. For convenience, we also transform the parameters for VI and MCMC so that $\boldsymbol{\vartheta} = \{\boldsymbol{\beta}, \log(\tau^2) \}$ for the ridge prior and $\boldsymbol{\vartheta} = \{\boldsymbol{\beta}, \log(\lambda_1^2), \log(\lambda_2^2), ..., \log(\lambda_p^2), \log(\tau) \}$ for the horseshoe prior.   We first introduce the HMC scheme in Section \ref{sec:MCMC}, before we develop the scalable VI approach in Section \ref{sec:VB}.

\subsection{Exact estimation using MCMC} \label{sec:MCMC}
Since the sampler of \cite{KleNotSmi2020} was too sticky to produce reliable estimates in our small data scenario, we suggest an alternative approach based on HMC to generate from $\varthetavec \mid \xvec, \yvec$.
HMC augments
$\varthetavec$ by momentum variables, and draws samples from an extended
target distribution that is proportional to the exponential of the
Hamiltonian function. The dynamics specify how the Hamiltonian
function evolves, and its volume-conserving property results in
high acceptance rates of the proposed iterates when tuned properly. To do so we employ the leapfrog integrator which involves the posterior $\log p(\varthetavec\mid\xvec, \yvec)$ and its gradient, see the Web Appendix A.1 for full details on HMC settings and the algorithm.

\subsection{Approximate estimation using VI}\label{sec:VB}
The general idea of VI is to turn sampling from the posterior into an optimization problem. In VI the posterior $p(\boldsymbol{\vartheta}\mid \xvec, \boldsymbol{y})$ is approximated by a member of some tractable density family $q_{{\lambdavec}}(\boldsymbol{\vartheta})$ that depends on a vector of variational parameters $\boldsymbol{\lambdavec}$. Proximity between $p(\boldsymbol{\vartheta}\mid \xvec, \boldsymbol{y})$ and $q_\lambdavec$ is measured by some measure of closeness. When this measure is the Kullback-Leibler divergence (KLD) from $q_\lambdavec$ to $p(\boldsymbol{\vartheta}\mid \xvec, \boldsymbol{y})$ 
\begin{equation*}
    \mbox{KLD}\left({q_{{\lambdavec}}(\boldsymbol{\vartheta})}\,||\,{p(\boldsymbol{\vartheta}\mid \xvec, \boldsymbol{y})}\right) = \int \log \Big(\frac{q_{{\lambdavec}}(\boldsymbol{\vartheta})}{p(\boldsymbol{\vartheta}\mid \xvec, \boldsymbol{y})} \Big) q_{{\lambdavec}}(\boldsymbol{\vartheta})d\boldsymbol{\vartheta},
    \end{equation*}
the optimal approximation maximizes the evidence lower bound given by \citep[ELBO;][]{OrmWan2010}
\begin{equation}\label{eq:Elb}
\mathcal{L}(\lambdavec)=\int q_{\lambdavec}(\varthetavec)\log \Big(\frac{p(\yvec,\varthetavec | \xvec)}{q_{\lambda}(\varthetavec)}\Big) d\varthetavec\,,
\end{equation}
with respect to $\lambdavec$. Setting $h(\varthetavec)=p(\yvec\mid \xvec,\varthetavec)p(\varthetavec)$, we note that \eqref{eq:Elb} is  an expectation with respect to $q_\lambdavec(\varthetavec)$, i.e.
\begin{equation}
\mathcal{L}(\lambdavec)=\dsE_{q_\lambdavec}[ \log h(\varthetavec)-\log q_{\lambdavec}(\varthetavec)],
\end{equation}
and this observation enables an unbiased Monte Carlo estimation of the gradient of $\mathcal{L}(\lambdavec)$ after differentiating under
the integral sign.  Doing so, the resulting expression for the gradient $\nabla_{\lambdavec}\mathcal{L}(\lambdavec)$ results  in an expectation with respect to $q_{\lambdavec}$,
\begin{equation}
\nabla_{\lambdavec}\mathcal{L}(\lambdavec)=\dsE_{q_{\lambdavec}}[\nabla_\lambdavec \log q_{\lambdavec}(\varthetavec)(\log h(\varthetavec)-\log q_{\lambdavec}(\varthetavec))]\,,
\end{equation}
where the so-called log-derivative trick $\dsE_{q_\lambdavec}[\nabla_\lambdavec\log q_{\lambdavec}(\varthetavec)]=0$ was used. This is often used with stochastic gradient ascent (SGA) methods to optimize the ELBO \citep[also known as stochastic VI, see e.g.~][]{NotTanVilKoh2012,HofBleWanPai2013,TitLaz2014}. 
 Denoting with $\reallywidehat{\nabla_{\lambdavec}\mathcal{L}(\lambdavec)}$  an unbiased Monte Carlo estimate of the gradient $\nabla_\lambdavec\mathcal{L}(\lambdavec)$ and with $\lambdavec^{(0)}$ an initial value for $\lambdavec$, SGA  performs the update
\begin{equation}\label{eq:SGA}
\lambdavec^{(t+1)}=\lambdavec^{(t)} + \rhovec^{(t)}\circ\reallywidehat{\nabla_{\lambdavec}\mathcal{L}(\lambdavec^{(t)})}\,
\end{equation}
recursively. In \eqref{eq:SGA}, $\circ$ denotes the Hadamard (element-by-element) product
of two random vectors and $\lbrace \rhovec^{(t)} \rbrace_{t\geq 0}$ 
is a sequence of vector-valued learning rates with dimension $\dim(\lambdavec)$. 
In practice, it is important to consider
adaptive learning rates, and we do so using the ADADELTA method of~\cite{Zei2012} as successfully adopted in~e.g.~\cite{OngNotSmi2017}.

\subsubsection{Choice for the variational approximation \texorpdfstring{$q_{\lambdavec}$}{TEXT}}
Successful application of stochastic VI  requires a numerically tractable yet flexible variational density $q_{\lambdavec}$. A popular choice for the approximation family is the Gaussian distribution, which contains the mean vector and the covariance elements as variational parameters. It has been shown to approximate posteriors well with similar copula regression models \citep{SmiKle2020}. To reduce computational cost, while still being flexible enough, we follow \citet{OngNotSmi2017} who use a factored representation of the covariance matrix. A typical member $q_{{\lambdavec}}(\boldsymbol{\vartheta})$ of this approximating family takes on the form
\begin{equation*} \label{eq:betterrepres}
    q_{\lambdavec}(\boldsymbol{\vartheta}) = N(\boldsymbol{\mu}, \boldsymbol{\Upsilon}\boldsymbol{\Upsilon}^T + \boldsymbol{\Delta}^2),
\end{equation*}
with $\boldsymbol{\Upsilon}$ being a lower triangular, full rank $p_\varthetavec \times k$ matrix with $k \ll p_\varthetavec$, and $\boldsymbol{\Delta}$ a real-valued diagonal matrix with diagonal elements $ \boldsymbol{d} = (d_1,..., d_{p_\varthetavec})$. The elements above the diagonal in $\boldsymbol{\Upsilon}$ are fixed to zero. Evidently, computing the elements of a $p_\varthetavec \times k$ matrix is much faster than computing the elements of the $p_\varthetavec \times p_\varthetavec$ covariance matrix directly. 

For this variational density, \citep{OngNotSmi2017} employ the re-parameterization trick \citep{KinWel2014,RezMohWie2014} to reduce variance of the Monte-Carlo estimates required for estimating the gradients unbiasedly. In our case this leads to re-writing the model parameters as $\varthetavec=\muvec+\mUpsilon\xivec+\dvec\circ\deltavec\coloneq g(\xivec,\deltavec)$,
where $\xivec\in\dsR^{k}$, $\deltavec\in\dsR^{p_{\varthetavec}}$ and $\xivec, \deltavec \sim N(\boldsymbol{0},\boldsymbol{I})$, $\lambdavec=(\muvec^\top,\mbox{vech}(\mUpsilon)^\top,\dvec^\top)^\top$ and vech($\mUpsilon$), stacks the elements of the lower triangle of $\mUpsilon$ (including the diagonal) columnwise
into a vector.

Evidently it is easy to sample from a standard normally distributed variable and we found $M=50$ iterates of (\xivec,\deltavec) for the gradient estimation to be sufficient. To implement the SGA \eqref{eq:SGA} we need the gradients
 $$\nabla_{\lambdavec}\mathcal{L}(\lambdavec)=(\nabla_{\muvec}\mathcal{L}(\lambdavec)^\top,
\nabla_{\mbox{\scriptsize{vech}}(\mUpsilon)}\mathcal{L}(\lambdavec)^\top,
\nabla_{\dvec}\mathcal{L}(\lambdavec)^\top)^\top$$ which involve the gradients of $h(\varthetavec)$ with respect to the elements of $\varthetavec$. The gradients can be obtained analytically, see the Web Appendix A.2.  The complete VI algorithm is summarized in Algorithm \ref{algo2}.

\begin{algorithm}[H]\captionsetup{labelfont={sc,bf}, labelsep=newline}
\caption{Variational approximation with a factored covariance structure}\label{algo2}
 \SetKwInOut{Input}{Input}
 \SetKwInOut{Output}{Output}
 \Input{Variational parameters $\boldsymbol{\lambda}^{(0)} = (\boldsymbol{\mu}^{(0)}, \mbox{\scriptsize{vech}}(\mUpsilon)^{(0)}, \boldsymbol{d}^{(0)})$, iteration $t= 0$}
 \Output{Optimal variational parameters $\hat{\boldsymbol{\lambda}}$}
 \While{ELBO has not converged}{
 Generate for $m=1,\ldots,M$ samples $\xivec^{(t, m)}, \deltavec^{(t, m)} \sim N(\boldsymbol{0},\boldsymbol{I})$. \\
 Compute the unbiased estimate of the gradient $\reallywidehat{\nabla_{{\lambdavec}} \mathcal{L}(\boldsymbol{\lambdavec}^{(t)})} \leftarrow (\reallywidehat{\nabla_{{\muvec}} \mathcal{L}(\boldsymbol{\lambdavec}^{(t)})}, \reallywidehat{\nabla_{\mbox{\scriptsize{vech}}(\mUpsilon)} \mathcal{L}(\boldsymbol{\lambdavec}^{(t)})}, \reallywidehat{\nabla_{\dvec} \mathcal{L}(\boldsymbol{\lambdavec}^{(t)})})$ using the $M$ samples $(\xivec^{(t, 1)}, \deltavec^{(t, 1)}),\ldots,(\xivec^{(t, M)}, \deltavec^{(t, M)})$.\\%using the $v$ samples. \\
 Update the learning rate $\boldsymbol{\rho}^{(t)}$ using ADADELTA.\\
 Update the variational parameters via $\lambdavec^{(t+1)} \leftarrow \lambdavec^{(t)} + \rhovec^{(t)} \circ \reallywidehat{\nabla_{{\lambdavec}} \mathcal{L}(\boldsymbol{\lambdavec}^{(t)})}$.
  $t \leftarrow t + 1$.
 }
\end{algorithm}

\section{Analysis in autonomous driving}\label{sec:results}
The basis for the IC-NLM is a deep neural network that maps from images of the ahead-lying street to the transformed steering angles (see Step 2 of Algorithm \ref{alg:fullinference}). We will first introduce the end-to-end learning architecture and details on the training process in Section~\ref{sec:architecture}. Afterwards, our analysis in autonomous driving consists of three parts. First, in Section~\ref{sec:accuracy}, we verify that VI is an accurate and scalable alternative to MCMC in the IC-NLM based on the two data scenarios introduced in Section \ref{sec:data}.  Second, in Section~\ref{sec:quality}, we benchmark the IC-NLM with two other competitive methods for obtaining predictive densities for end-to-end learners with regard to calibration, coverage rates of the prediction intervals and the ability to identify high-risk predictions. Third, we explore how predictive densities for the steering angle can help to enhance the explainability of end-to-end learners in Section~\ref{sec:expl}. 

\subsection{End-to-end learner architecture and training} \label{sec:architecture}
\begin{figure}[htbp]
\includegraphics[scale = 0.33]{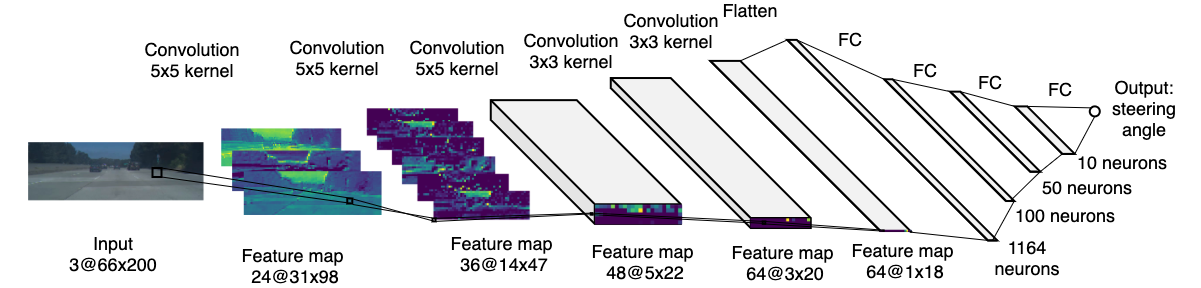}
\caption{PilotNet architecture by \cite{BojTesDwo2016} with activations from the comma2k19 driving data.}
\label{pilotnet}
\end{figure}

As an end-to-end learner we use a CNN based on the pioneering PilotNet architecture by \cite{BojTesDwo2016} with additional regularization to prevent overfitting. We deliberately decide against exploring more complex network architectures (such as LSTMs) to ensure comparability with the literature. Figure \ref{pilotnet} depicts the PilotNet architecture. 
The PilotNet is a CNN with 5 convolutional layers followed by 4 fully connected layers. CNNs are particularly suitable to deal with image data since they possess a ``weight-sharing'' property. This allows CNNs to learn features independently of their position in an image and reduces the number of training parameters. Each convolutional layer is followed by batch normalization and we use dropout on the fully connected layers to avoid overfitting. We use ReLu activation functions for all layers but the last layer which has a linear activation function and dimension $p_\varthetavec = 10$.  For Step 2 of Algorithm \ref{alg:icnlm} the networks are trained on the transformed steering angles $z_i=\Phi^{-1}(\hat F_Y(y_i))$ to minimize the MSE with the Adam optimizer. Details on training settings and the preprocessing steps to obtain a good model  can be found in the Web Appendix B.1.

To estimate the marginal density of the steering angle in Step 1 of Algorithm \ref{alg:fullinference}, we use a  non-parametric KDE with  Gaussian kernel \citep{Rac2008}. We found that this estimator was able to capture the shape of the density sufficiently well, while producing no numerical issues on the edges of the data distributions where observations are sparse. Driving on highways involves relatively little steering action. Most steering angles therefore lie in the interval of $[-15^{\circ}, +15^{\circ}]$ and we limit the steering angles to $[-50^{\circ}, +50^{\circ}]$. Angles outside of this interval are associated with complex steering actions such as U-turns or parking. These actions are beyond the scope of present-day end-to-end learners. 
We fit the IC-NLM to the end-to-end learner trained on both data sets once using both the ridge and horseshoe prior.

\subsection{Estimation accuracy of VI vs. HMC} \label{sec:accuracy}
\begin{figure}[h!] 
    \centering \textbf{Convergence and estimation accuracy of VI vs.~HMC using the small data with the ridge prior} \\
    \vspace{-0.25cm}
    \centering 
    \subfloat[]{{\includegraphics[height=3.5cm]{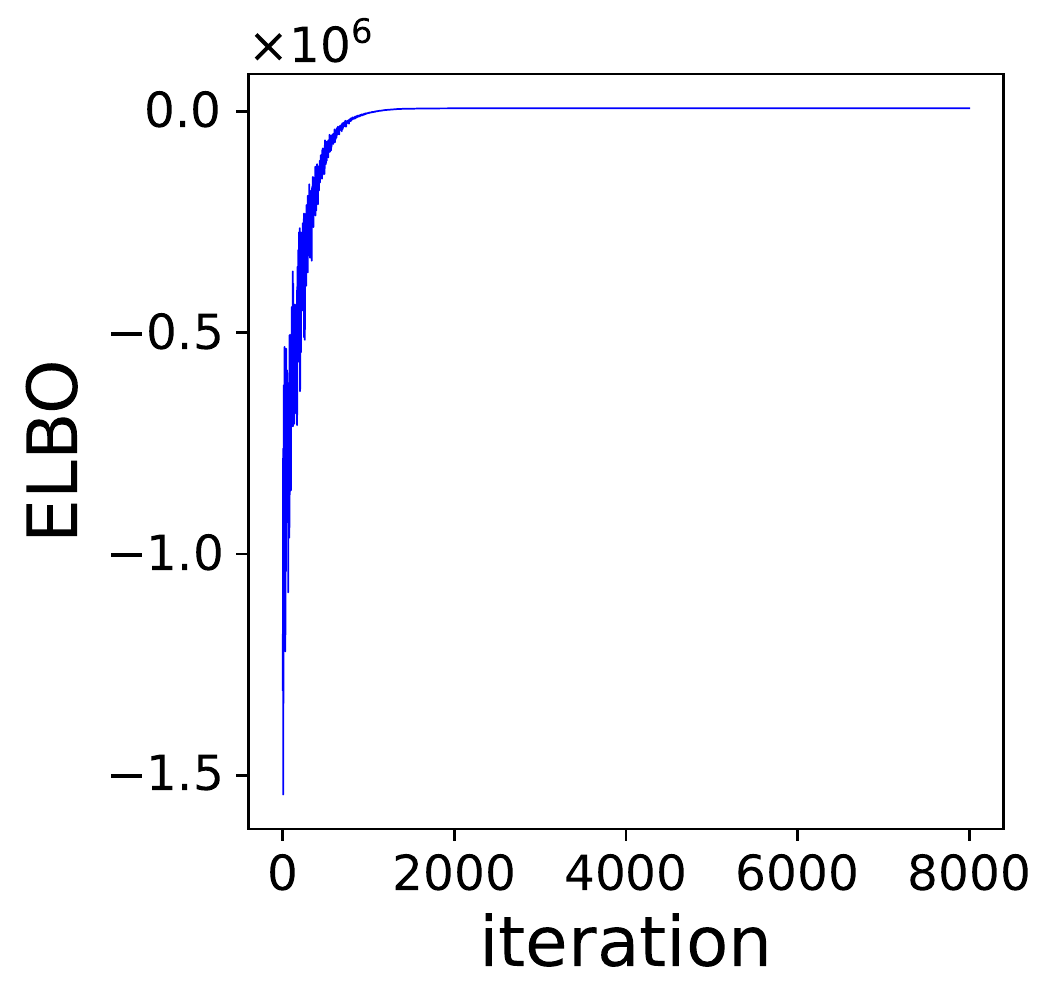} }} \hspace{-0.15cm}
    \subfloat[]{{\includegraphics[height=3.3cm]{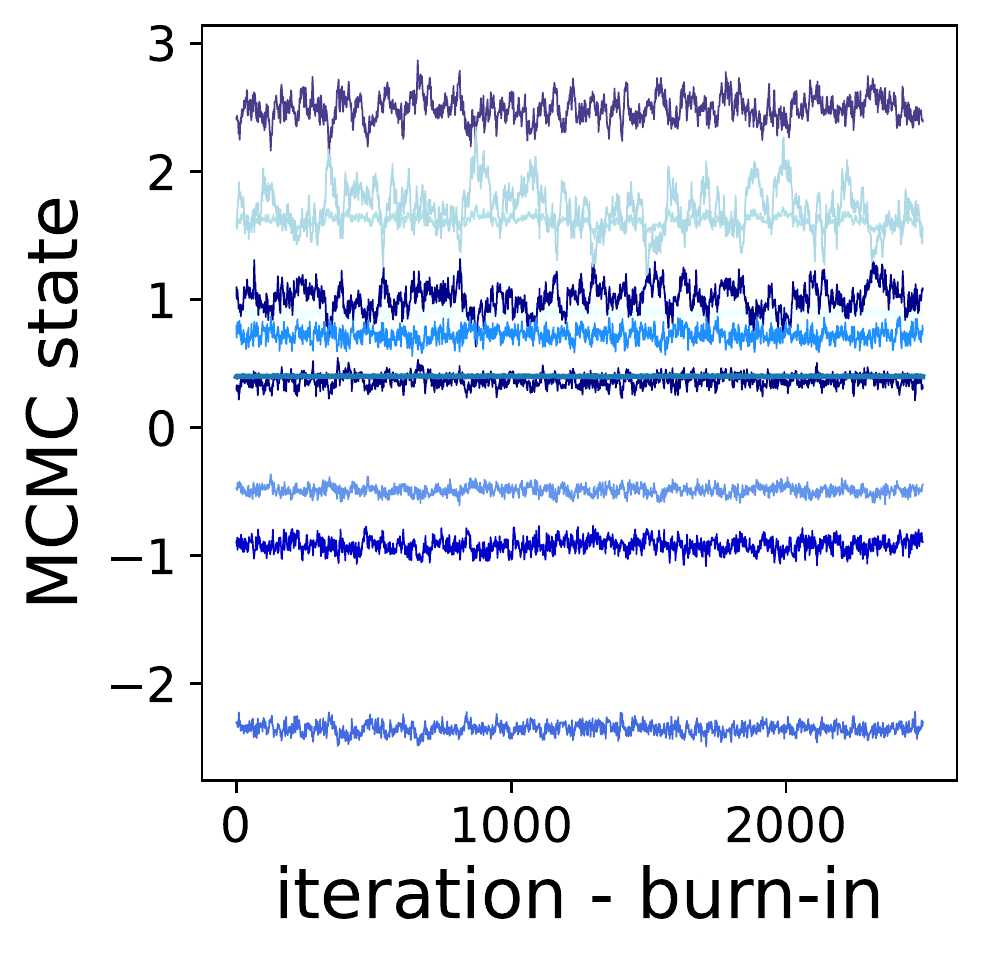} }}
    \subfloat[]{{\includegraphics[height=3.3cm]{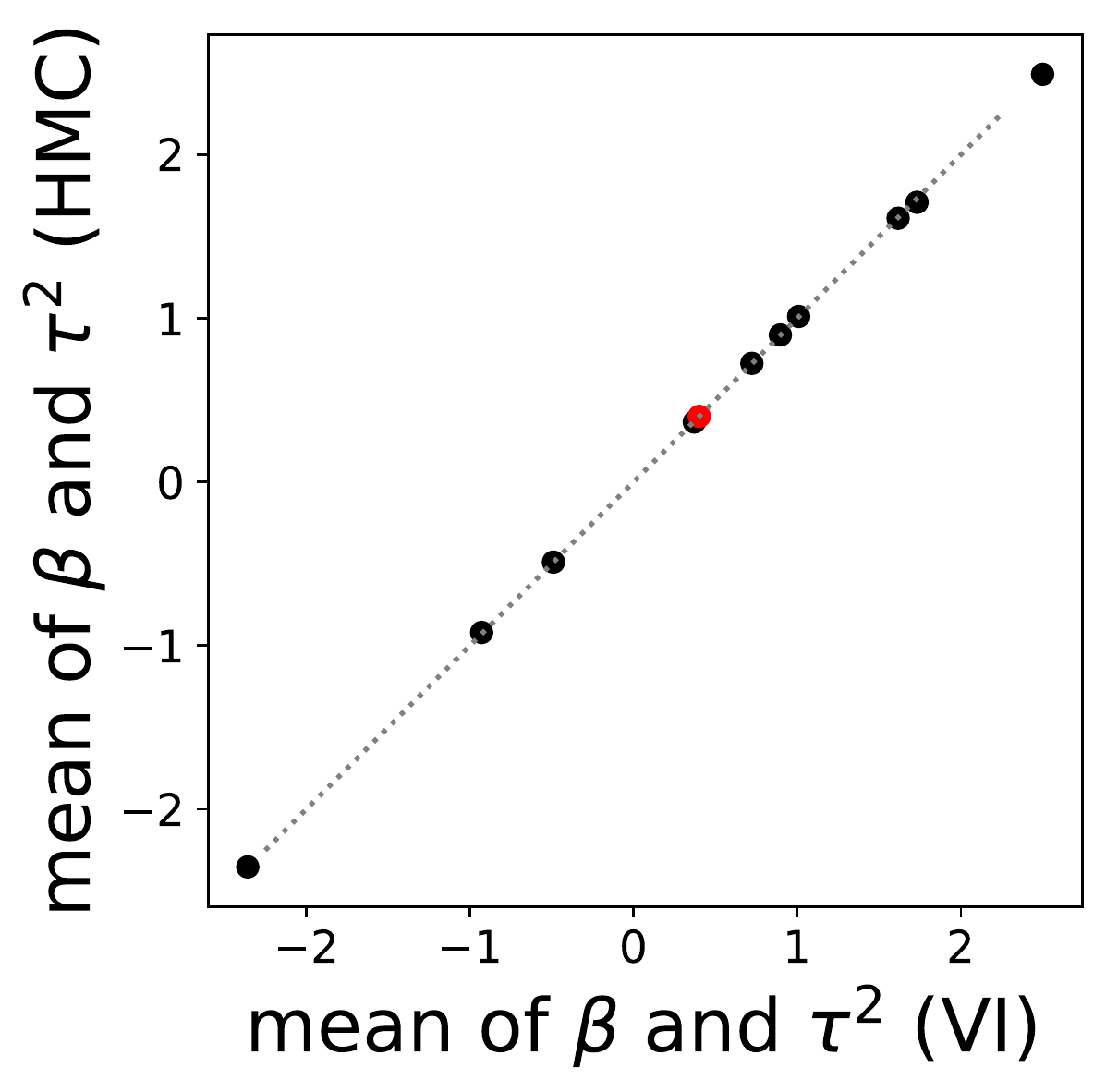} }}
    \subfloat[]{{\includegraphics[height=3.3cm]{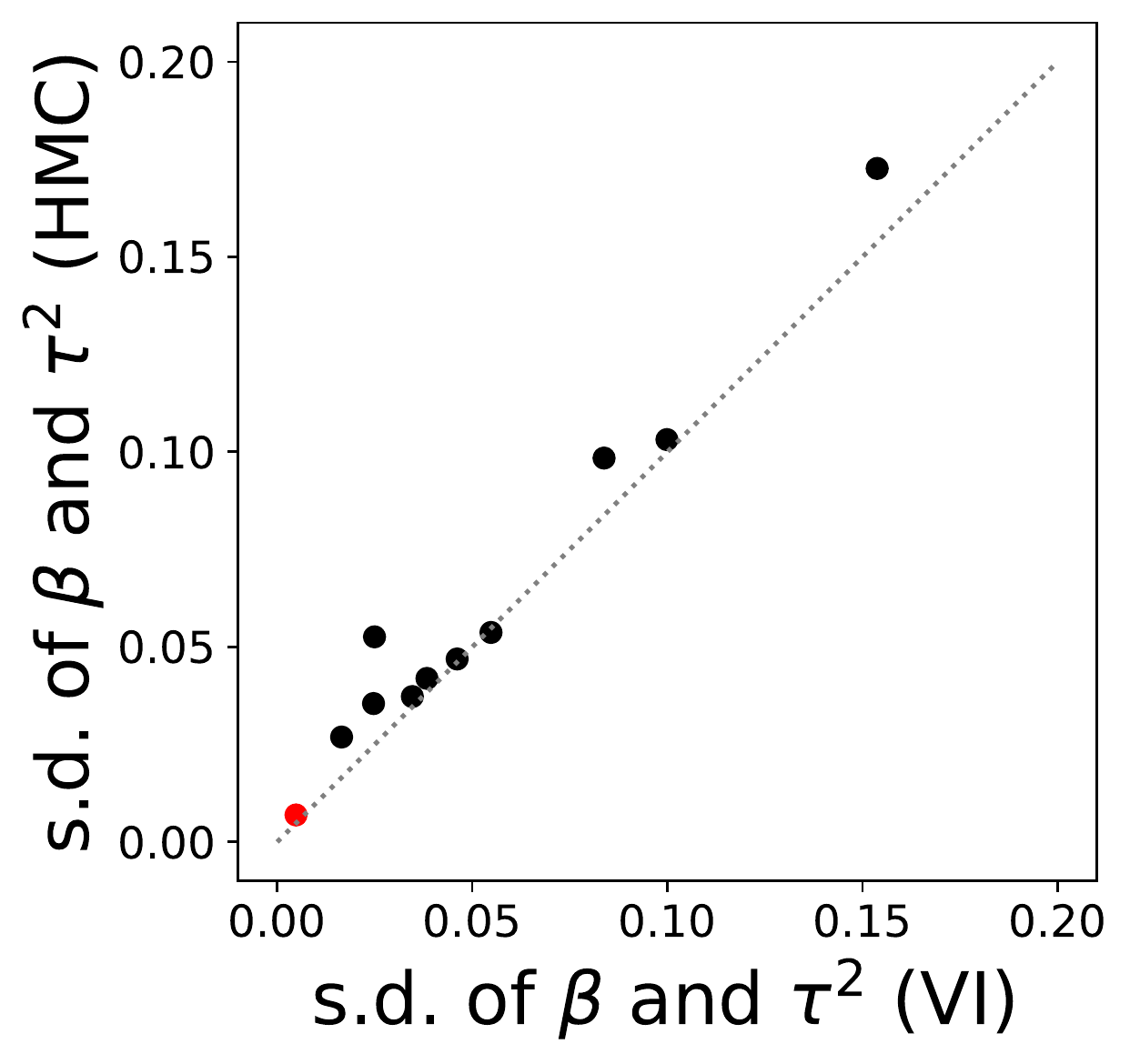} }} \\
    \caption{Panel (a) shows the ELBO for VI with $k=3$ factors. Panel (b) shows trace plots of HMC, where the x-axis denotes the number of iterations minus the number of burn-in samples. Panels (c) and (d) show posterior means and standard deviations of $\boldsymbol{\beta}$ (black) and $\boldsymbol{\theta} = \{\tau^2\}$ (red) for VI and HMC.}
    \label{fig:CPL_acc}
\end{figure}

\begin{figure}[ht]%
    \centering \textbf{Convergence and estimation accuracy of VI vs.~HMC using the small data with the horseshoe prior} \\
    \vspace{-0.25cm}
    \subfloat[]{{\includegraphics[width=3.5cm]{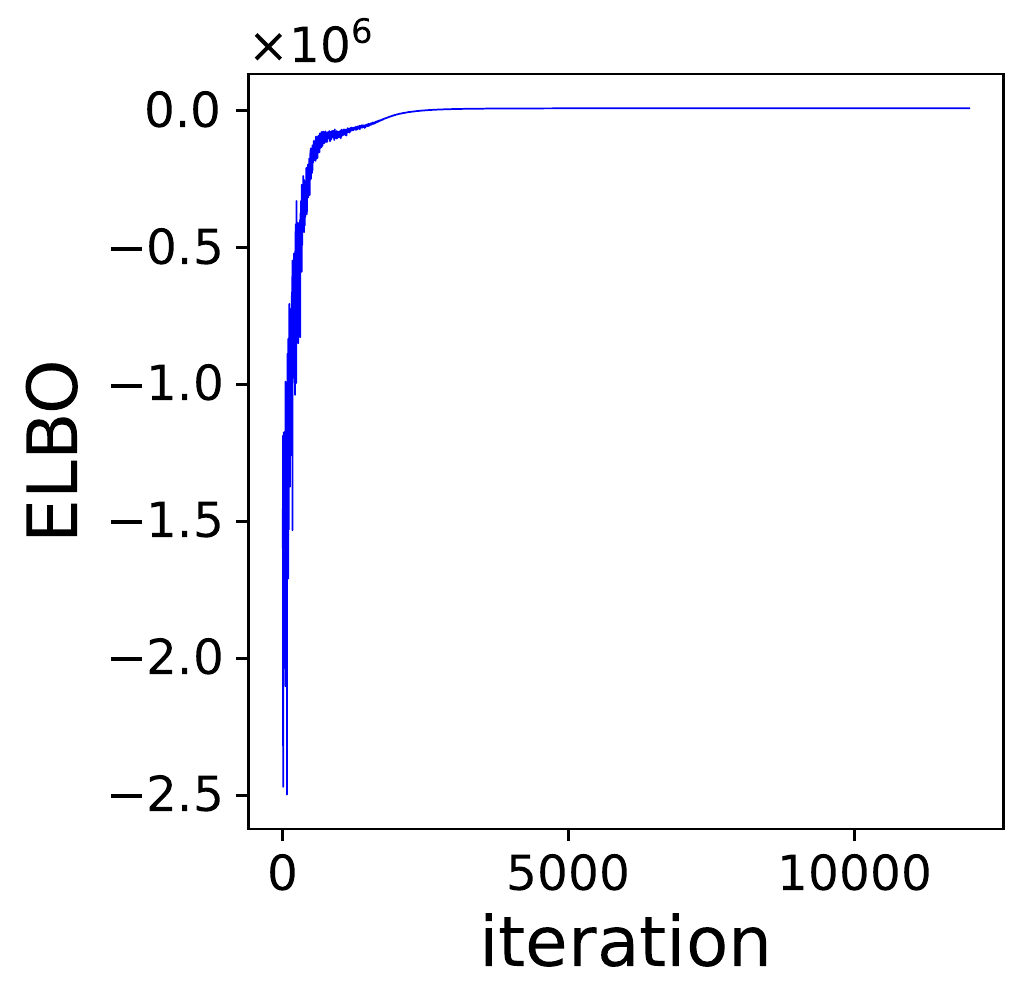} }}  \hspace{-0.15cm}
    \subfloat[]{{\includegraphics[width=3.3cm]{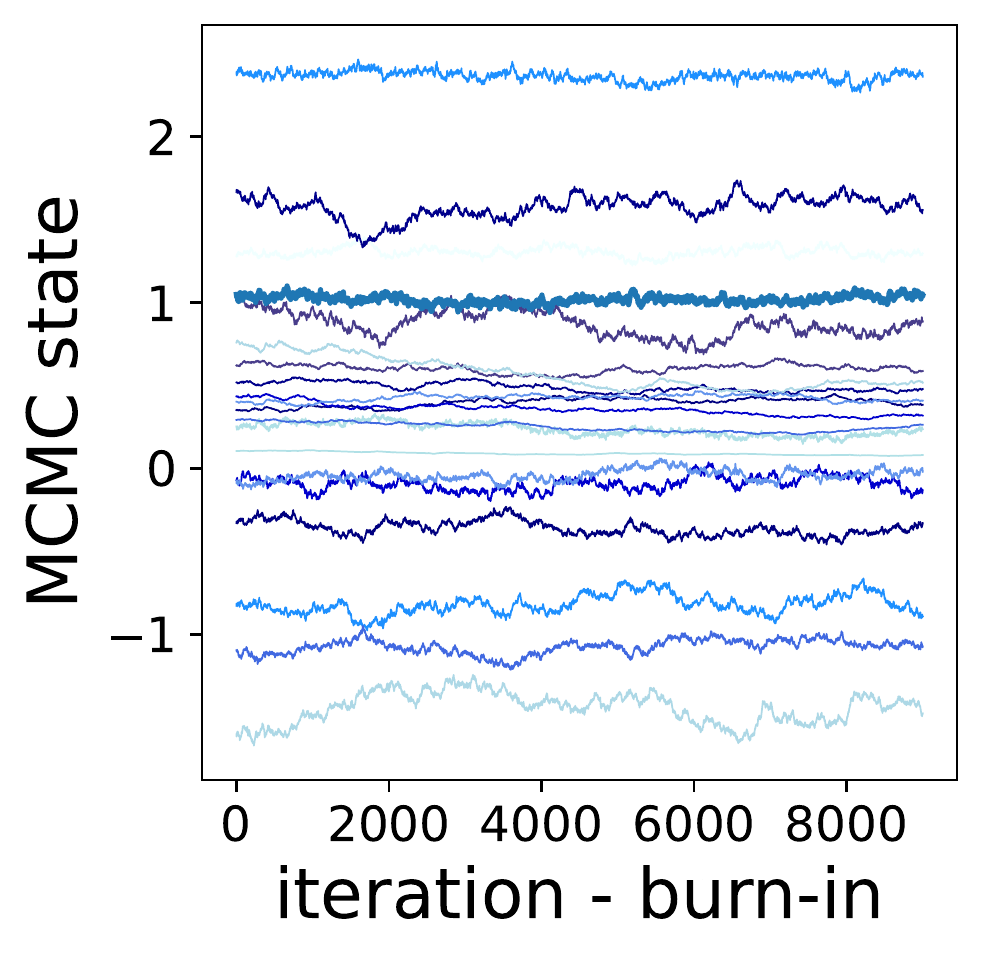} }}
    \subfloat[]{{\includegraphics[width=3.3cm]{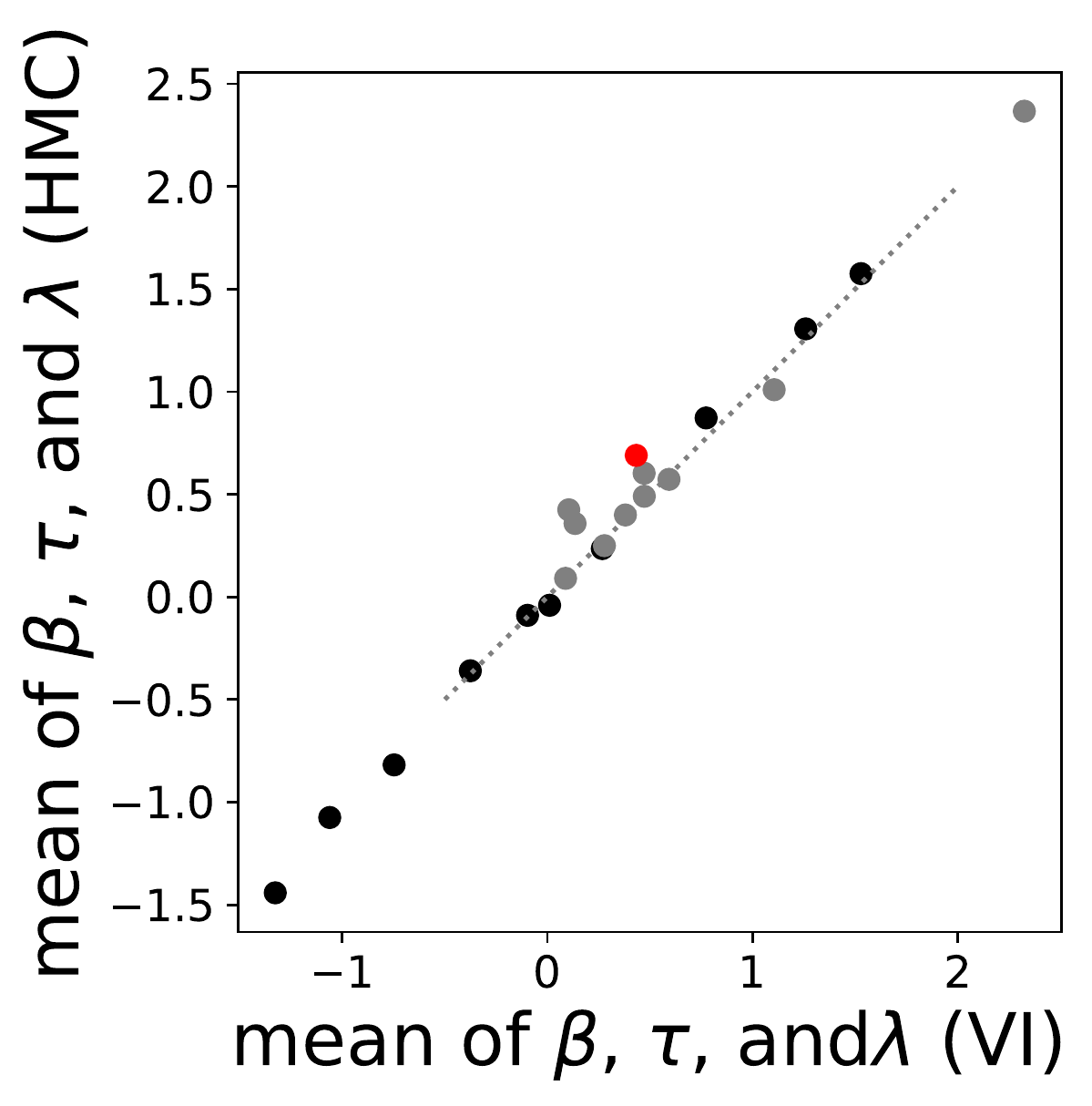} }}
    \subfloat[]{{\includegraphics[width=3.3cm]{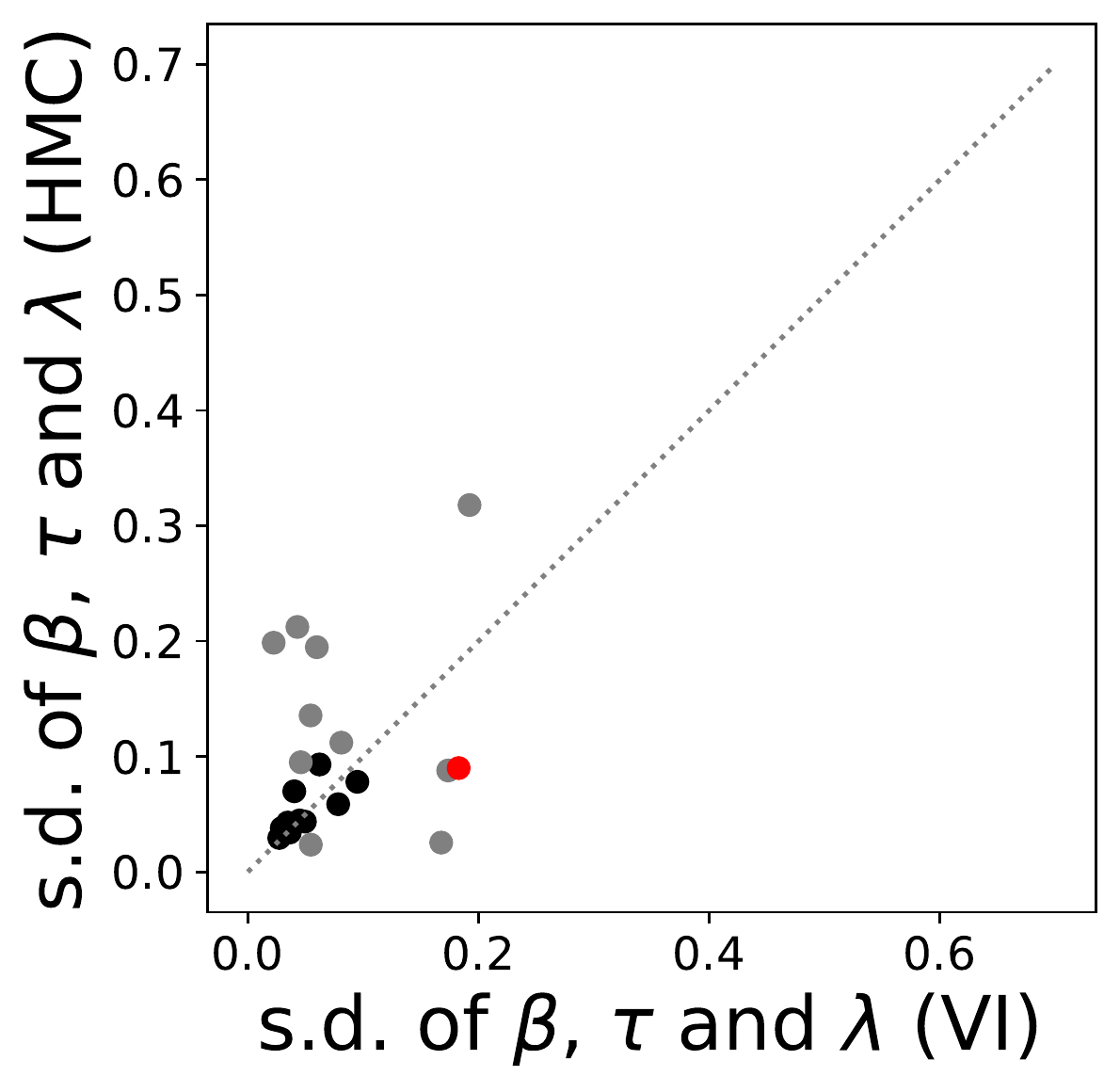} }} \\
    \caption[Accuracy of VI ($k$ = 3) vs. HMC for the horseshoe prior]{Panel (a) shows the ELBO for VI with $k=3$ factors. Panel (b) shows trace plots of HMC. Panels (c) and (d) show posterior means and standard deviations of $\boldsymbol{\beta}$ (black), $\boldsymbol{\lambda}$ (grey) and $\tau$ (red).}
   \label{fig:CPL_acc_horse}
\end{figure}

To assess the pure estimation accuracy of VI across the two data scenarios and priors, we compare the VI results with those from HMC. 
Figure~\ref{fig:CPL_acc} shows the results for the small data scenario using a ridge prior. Panels (a) and (b) show the lower bound from VI with $k=3$ factors and trace plots from HMC, respectively. Posterior means and standard deviations of model parameters can be found in panels (c) and (d), respectively. In these two right hand plots, points on the diagonal indicate that the respective posterior means or standard deviations using VI and HMC coincide. Figure~\ref{fig:CPL_acc_horse} presents the same results but for the horseshoe prior.

With both priors, VI converges fast, while HMC chains  exhibit obvious autocorrelations when the horseshoe prior is used. For both priors, VI estimates the posterior means very accurately, while the standard deviations suffer from slight over- and underestimation only in a  very few cases with a ridge prior. When using a horseshoe prior, deviations in standard deviations are more pronounced. This over- and underestimation did not translate to deviations between the final predictive densities for the steering angles obtained by VI and HMC as can be seen when inspecting the predictive accuracy in the following section. That is why we adopted no further measures to improve the estimation of the standard deviations.
 Predictive densities and predictive accuracy is improved under the more complex shrinkage prior, as we will see in Section \ref{sec:quality}. The results for the large data scenario are presented in the Web Appendix C and are similar to those for the small data scenario.
In terms of runtime, VI is much faster than HMC. This becomes more evident in the computationally more intensive case with the horseshoe prior and in particular for the large data scenario, where using VI reduces the computation time by several days when compared to HMC (both using two 12-core CPUs). All runtimes are reported in Appendix B.2.

\subsection{Benchmark study} \label{sec:quality}
Having ensured the scalability and accuracy of the VI approach, we will now benchmark the predictive accuracy, calibration and uncertainty estimates obtained by the IC-NLM. We compare the performance on each of the two validation sets and both priors with two non-probabilistic and two probabilistic end-to-end learners. Overall, we compare the following models
\begin{itemize}
\item \textbf{Naive learner}: Driving model that always drives straight, regardless of the input.
\item \textbf{Uncalibrated learner}: PilotNet trained directly on the response $\boldsymbol{y}$.
\item \textbf{IC-NLM with ridge/horseshoe prior}: IC-NLM based on the PilotNet trained to predict \\ $z_i = \Phi_1^{-1}(\hat{F}_Y(y_i))$ as described in Algorithm \ref{alg:fullinference}. We fit two versions, one using VI and one using HMC.
\item \textbf{MC dropout}: PilotNet model using Monte Carlo (MC) dropout \citep{GalGha2016} with dropout probabilities from \cite{MicKwiGal2018}, and using 1{,}000 dropout masks at prediction time. 
\item \textbf{Mixture density network (MDN)}: Mixture density network \citep{Bis1994} with 50 mixture components. An MDN is a DNN where the output is a Gaussian mixture and the model is trained to predict the means, variances and weights of the mixture components by maximizing the likelihood of the response. 
\end{itemize} 
All models are implemented using TensorFlow \citep{Aba2015} and Keras \citep{Cho2015}. We implement MC dropout ourselves and for the MDN we use the Keras-compatible keras-mdn-layer \citep{MarDuh2020}.
Note that the first two models do not produce predictive densities, so they will only be used in the comparison of the predictive point accuracy which we will measure via the mean absolute error (MAE) and the corresponding mean squared error (MSE) on the validation set. In order to compare predictive densities for the IC-NLM, MC-Dropout and MDN, we follow the literature and investigate the deviation between the true and mean predictive angle. Specifically in the results,  we refer to Accuracy I  as the percentage of predictions, for which the true and predicted steering angle differ by less than six degrees. Related to this, Accuracy II measures the percentage of steering angles where the predictive and true value differ by less or exactly two degrees. 

Results for these measures are presented first, before we investigate the calibration and uncertainty estimates  in more detail. Since the general tendency of results is similar for both data scenarios, we present the ones for the small data scenario in the following and refer the reader to the Web Appendix C for respective results in the large scenario. 

\paragraph*{Predictive accuracy} 
 Table~\ref{tab:performance}  shows the MAE, MSE, Accuracy I and Accuracy II as defined above for all methods. 
\begin{table}[ht] 
\caption{Benchmark study. Predictive performance in the small data scenario as measured by the MAE, MSE Accuracy I and Accuracy II (from left to the right) on the validation set. The first column details the employed models.}
\label{tab:performance}
\centering
\begin{tabular}{ c c c c  c } 
\hline
 Model &  MAE & MSE &  Accuracy I (\%) & Accuracy II  (\%)   \\ 
 \hline
 Naive learner & 2.16 & 14.61 &  93.21 & 65.12   \\ 
 Deterministic DNN &  1.42 & 6.43 &  98.19 & 79.61  \\ 
  IC-NLM + ridge + HMC  & 1.34 & 6.28 &  98.08 & 82.22   \\
  IC-NLM + ridge + VI   & 1.34 & 6.28 &  98.08 & 82.21  \\
  IC-NLM + horseshoe + HMC  & \textbf{1.29} & {5.67} &  98.21 & \textbf{82.75}   \\
  IC-NLM + horseshoe + VI  & 1.29 & \textbf{5.66} &  98.21 & 82.73    \\
 MC-Dropout & 1.41 & 6.35 & \textbf{98.37} & 80.08  \\
 MDN & 1.36 & 6.63 & 98.16 & 81.91 \\
 \hline
 \end{tabular}
\end{table}
The IC-NLM using a horseshoe prior performs best on all metrics but on Accuracy I, where MC-Dropout performs best. Performance for VI is very similar to the one of HMC which again underpins the good accuracy of our approximate scalable method. 
The choice of the prior for the IC-NLM is relevant and the more complex horseshoe prior leads to more accurate point predictions. Hence it is worth the additional computational cost. 

\paragraph*{Calibration} 
 
\begin{figure}[ht]%
    \centering \textbf{Marginal and probabilistic calibration in the small data scenario} \\
    \subfloat[Marginal calibration]{{\includegraphics[width=7cm]{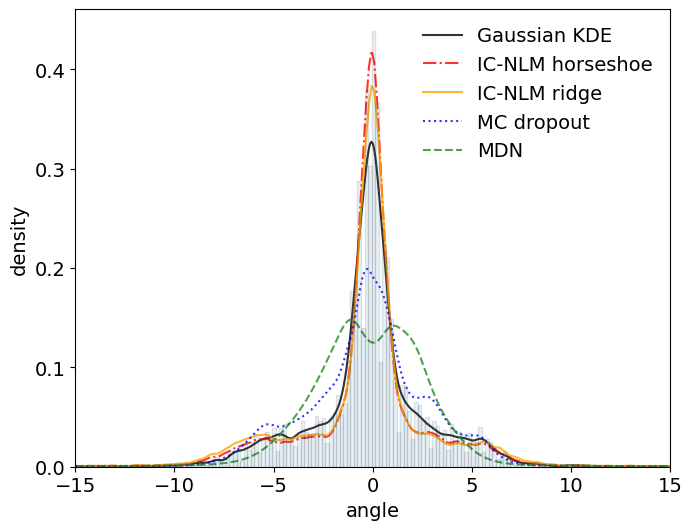}}}
    \hspace{0.18cm}
    \subfloat[Probabilistic calibration]{{\includegraphics[width=7cm]{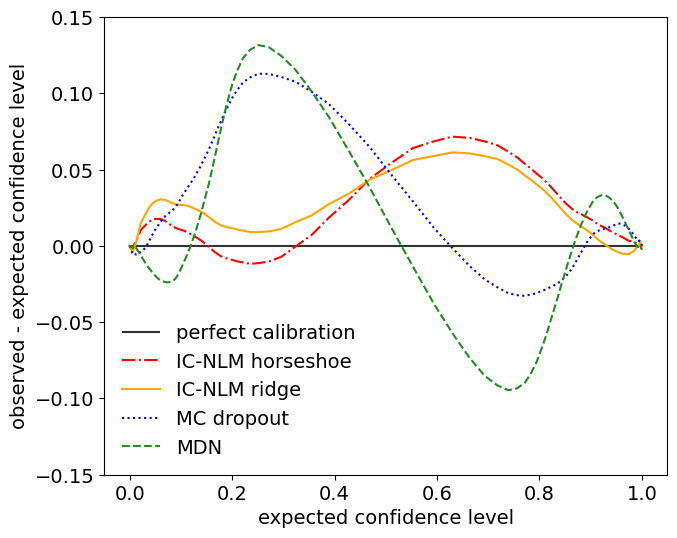}}} 
    \caption{Benchmark study. Marginal (panel (a)) and probabilistic (panel (b)) calibration for the validation data with the IC-NLM with ridge (solid yellow line), IC-NLM with horseshoe (dot-dashed red line), MC-Dropout (dotted blue line), and MDN (dashed green line).}
   \label{fig:CPL_marg_cal}
\end{figure}

Calibration is essential to the reliability of predictive densities and one main motivation for the IC-NLM is to achieve marginal calibration (see Section~\ref{sec:calibration}).
Figure~\ref{fig:CPL_marg_cal} (a) shows the marginal calibration plots of all probabilistic learners for the validation data along with the histogram and the KDE. For the IC-NLM we show the results from VI only, since they are visually indistinguishable from the HMC results. Accurate marginal calibration occurs when the predictive densities coincide or are close to the respective KDE. Even though  the IC-NLM is by construction only marginally calibrated for the training data, it shows the most accurate marginal calibration also for the validation set. In contrast,  MC-Dropout and  MDN are not marginally calibrated. 
\\
Figure~\ref{fig:CPL_marg_cal} (b) shows the probabilistic calibration of the end-to-end learners (see Section~\ref{sec:calibration}). Probabilistic calibration is measured by the difference between the observed and expected confidence level plotted over the confidence level. Even though probabilistic calibration is not guaranteed with the IC-NLM, compared to the two benchmarks MC-Dropout and  MDN it indeed performs best for both priors as can be seen in Figure~\ref{fig:CPL_marg_cal} (b).

\begin{figure}[ht]%
    \centering \textbf{Coverage rates and  of high-error predictions in the small data scenario} \\
    \subfloat[Prediction interval reliability]{{\includegraphics[width=7cm]{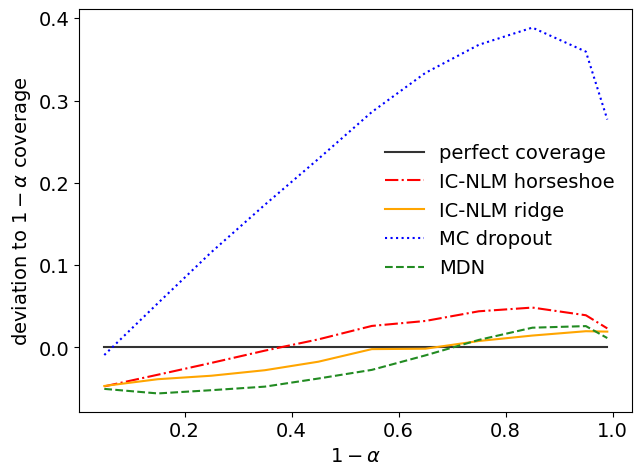}}} \hspace{0.18cm}
    \subfloat[Prediction error vs. predictive variance]{{\includegraphics[width=7cm]{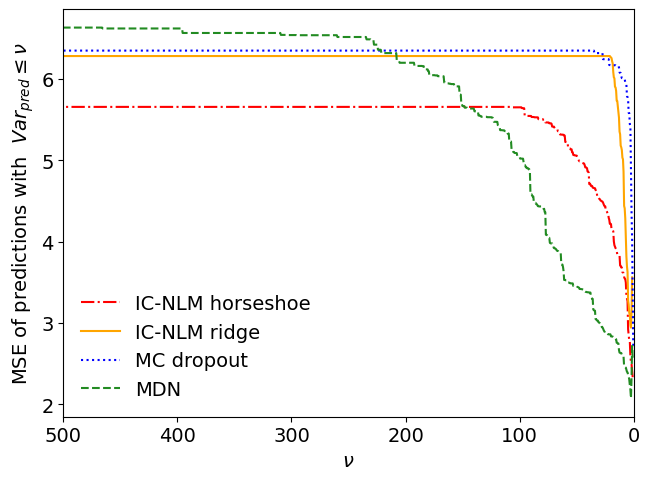}}}
    \caption{Benchmark study.  Coverage rates and error-variance relation for the validation data with the IC-NLM with ridge (solid yellow line), IC-NLM with horseshoe (dot-dashed red line), MC-Dropout (dotted blue line), and MDN (dashed green line).}
   \label{fig:PI_coverage_rates}
\end{figure}

\paragraph*{Prediction intervals, coverage rates and high-error predictions}
Predictive densities should not only be calibrated but also provide reliable and ideally sharp prediction intervals. They should be reliable in the sense that their coverage rates are accurate and sharp in the sense that given accurate the width of prediction intervals is tight. The first property can be measured through the deviation between the expected and observed coverage rates of the prediction intervals over the confidence level as shown in Figure~\ref{fig:PI_coverage_rates} (a). For the small scenario, the IC-NLM with a ridge prior shows the most accurate prediction intervals, closely followed by the MDN and IC-NLM with a horseshoe prior. MC-Dropout produces unreliable prediction intervals, likely due to the usage of only 1{,}000 dropout masks at prediction time\footnote{Using more dropout masks, however, is associated with increasing computational costs and is unlikely to be an efficient solution in the context of autonomous driving.}.\\
Another usage of predictive densities is to identify overconfident learners.
Optimally, high variances should point to predictions that are associated with high errors, such that a high predicted variance can act as an early-warning for wrong predictions \citep{HeLakYee2020}. Predictive densities can then be used to identify predictions that are potentially wrong and initiate a switch to human steering. In overconfident learners the opposite relation holds and these learners can be dangerous when deployed in real road traffic. Overconfident learners can be identified by plotting the MSE for all observations carrying a predictive variance under a certain threshold $\nu$ against this threshold as done in Figure~\ref{fig:PI_coverage_rates} (b). A perfect uncertainty quantifier would produce a line from the lower right corner to the upper left corner, so that the prediction error increases linearly as the predicted variance increases. Figure~\ref{fig:PI_coverage_rates} (b) shows that no end-to-end learner exhibits this perfect relation between predictive variance and predictive error. But the MDN comes quite close and is by far the most reliable uncertainty quantifier, followed by the IC-NLM with a horseshoe prior. For MC-Dropout and the IC-NLM with a ridge prior there exists almost no relation between prediction error and predictive variance. This indicates that both of these learners are quite overconfident. 

Overall, the IC-NLM performs best in terms of point predictions, calibration and accuracy of the coverage rates. Only for the relation between predictive variance and predictive error the MDN performs better. This shows that the IC-NLM is a competitive option to obtain reliable predictive densities for the steering angle in autonomous driving applications.

\subsection{Understanding end-to-end learners}\label{sec:expl}
End-to-end learners generally are black-box models, which means that their predictions cannot easily be explained. In the context of autonomous driving, it is desirable to know how much of its surroundings an end-to-end learner actually understands, so that its safety can be precisely assessed. Predictive densities can be used to gain better insight on how an end-to-end learner sees its surroundings. A well-suited scenario to check whether an end-to-end learner understands its environment are situations where multiple steering actions are valid, e.g.~at intersections. In this case one can check whether the predictive density for the steering angle has several modes, where each mode should correspond to a valid steering actions \citep[compare e.g.~][]{XuGaoYuTre2016}. This is also important for combining end-to-end learners with route planning. Beyond that, we are often interested in the behavior of end-to-end learners in new environments, e.g.~novel lane patterns or road conditions that were not part of the training set. 

\begin{figure}[ht]%
    \centering \textbf{Identification of multiple steering options (small scenario, training)}
    \subfloat{{\includegraphics[width=7cm]{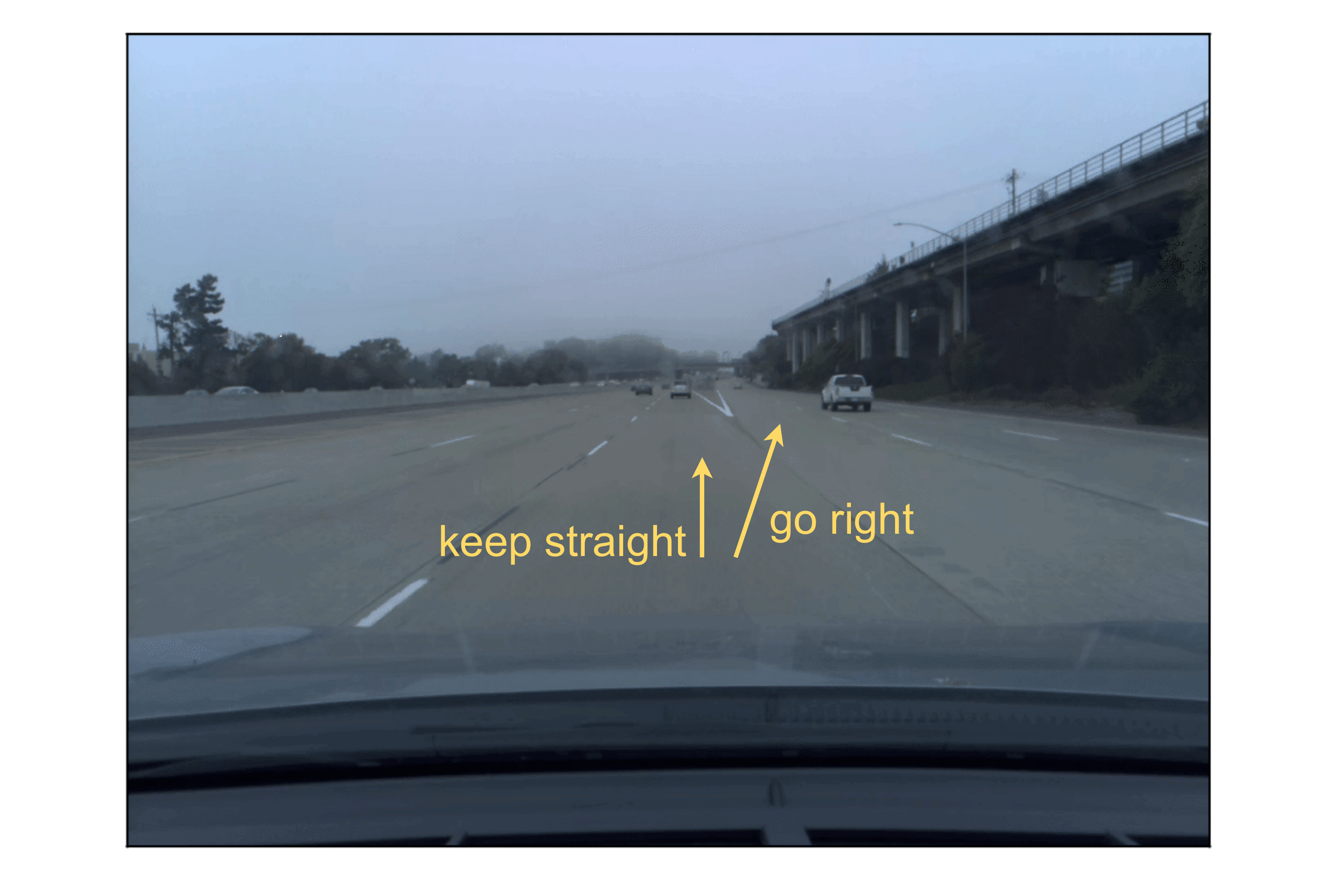}}}
    \subfloat{\raisebox{-.3cm}[0pt][0pt]{\includegraphics[width=6.5cm]{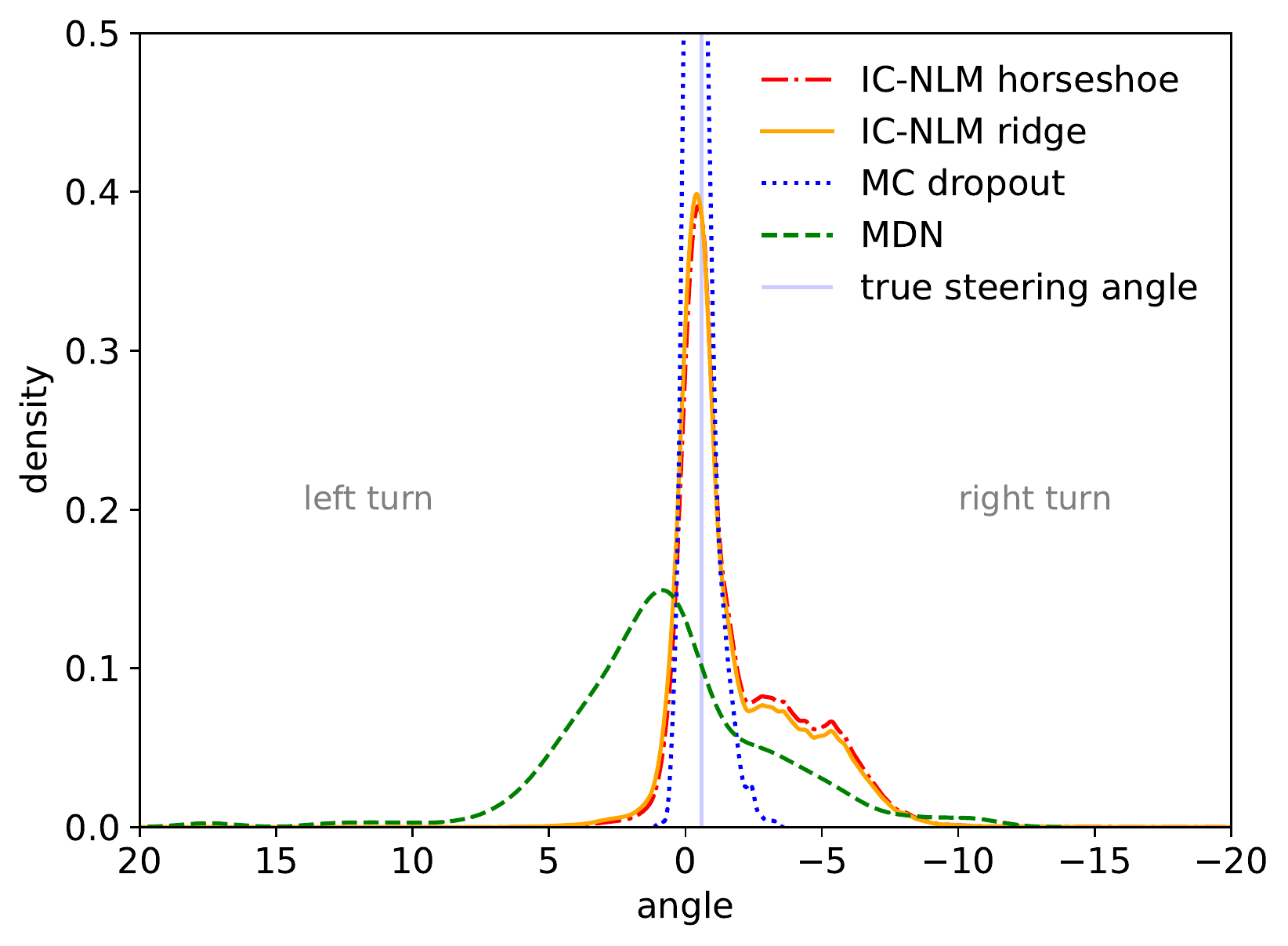}}} \\
    \vspace{.5cm}
    \centering \textbf{Uncertainty in the presence of an unknown lane pattern (large scenario, validation)} \\
    \subfloat{{\includegraphics[width=7cm]{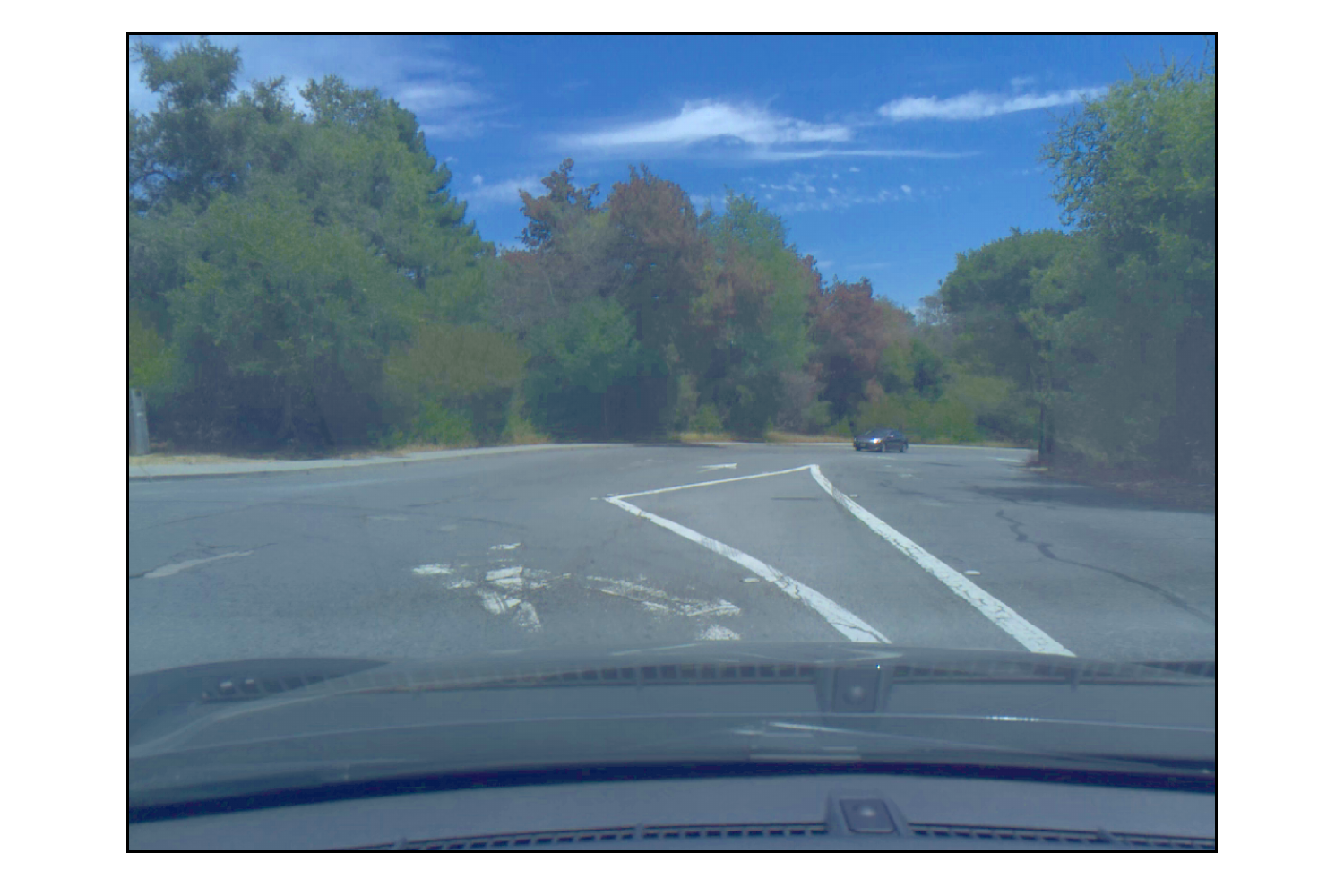}}} \centering
    \subfloat{\raisebox{-.3cm}[0pt][0pt]{\includegraphics[width=6.5cm]{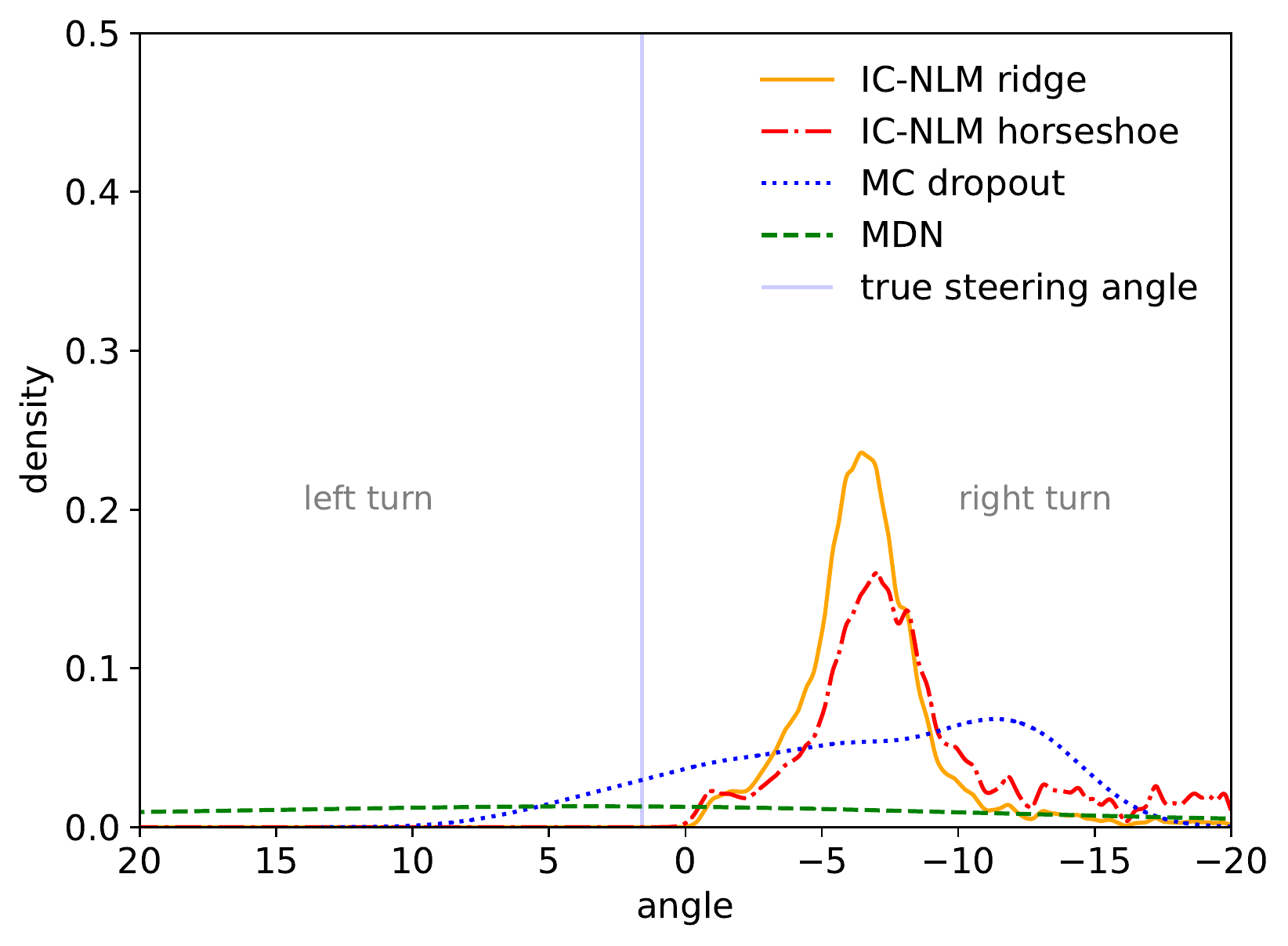}}} \\
    \caption{Two exemplary predictive densities. Top row: for the small, cleaned data scenario and an image from the training set. Bottom row: large, raw data scenario with an image from the validation set. Shown are IC-NLM with ridge (solid yellow line), IC-NLM with horseshoe (dot-dashed red line), MC-Dropout (dotted blue line), and MDN (dashed green line) along with the true steering angle (grey). }
   \label{fig:potentialsteer}
\end{figure}

Figure~\ref{fig:potentialsteer} shows two example images of the comma2k19 data with the predictive densities from the probabilistic end-to-end learners. The upper image features a parting road with two valid steering options (keep straight or go right) in the large training set. The lower image contains a distribution-shifted road pattern in the large validation set, which allows to explore how the end-to-end learners react in situations of high uncertainty. 
In the top row of Figure~\ref{fig:potentialsteer}, the two IC-NLM models manage to assign probability mass to the alternative steering action. MC-Dropout and MDN, however, both deliver almost unimodal predictions. Hence, MC-Dropout and MDN seem to have a limited understanding of the meaning of a parting lane and fail to indicate a second valid steering action. \\
In the bottom row of Figure~\ref{fig:potentialsteer} an unknown lane pattern appears that was not part of the training set. All probabilistic learners correctly associate high uncertainty with this image and are not overconfident. Especially the MDN does not favor any particular steering angle. 

These examples also illustrate that it is not irrelevant which probabilistic learner is used since for a single image the predictive densities of the learners can differ notably.  
\newpage

\section{Discussion}\label{sec:discussion}
We have expanded the IC-NLM of \cite{KleNotSmi2020} to a scalable version using VI to enable fast uncertainty quantification for the steering angle with end-to-end learners. A  detailed case study using the comma2k19 highway driving data set shows that the proposed VI approach is as accurate as exact inference based on MCMC and much faster in large $n$ scenarios. The scalable IC-NLM is competitive with other uncertainty quantification methods, namely MC-Dropout and MDNs. In terms of calibration, the IC-NLM even is superior. The choice of the prior in the IC-NLM has an impact on the predictive performance and reliable uncertainty quantification. Thus, we recommend to use priors with more complex shrinkage schemes, even though they generally lead to longer runtimes. Priors that allow for even more flexibility could potentially improve calibration further. In fact, one disadvantage of our recommended prior, the horseshoe prior, is that the shrinkage of larger coefficients can be hard to control, thus resulting in potential undershrinking of large coefficients \citep{PiiVeh2017}. The regularized horseshoe prior \citep{PiiVeh2017} solves this problem by shrinking the larger coefficients like a slab and the smaller coefficients like a horseshoe prior. Future research could investigate whether better calibration can be achieved by the regularized horseshoe. Beyond that, estimating the posterior with VI could be integrated into the loss function by maximizing the likelihood to speed up the estimation process as is done in other NLMs \citep{SnoRipSwe2015, RiqTucSno2018,ObeRas2019,PinGorNal2019}. 

Future research could also address the over- and underestimation of the standard deviations in the posterior of the model parameters when using VI. A possible reason for this problem could be gradient noise, which also translates to noise in the ELBO. For further improving the accuracy of the standard deviations, one could adapt methods in the spirit of \citet{MilFotDamAda2017}, who introduce a control variate to reduce gradient noise in VI and reach better convergence without having to increase the number of  samples $M$ in the re-parameterization trick to very large numbers. 

Another relevant research question in the future is to quantify the information loss by performing Bayesian inference only in the last layer of the DNN, compared to Bayesian inference on the weights in all layers. As has been shown in this work, the posterior densities of the NLM provide useful and comprehensible information about predictive uncertainty. Nonetheless, the negligence of uncertainty of all preceding layers should be investigated further also with respect to computational feasibility for large $n$ scenarios. A promising route could be to build on the fully Bayesian framework of deep generalized mixed models of \cite{TraNguNotKoh2020}.

Moreover, in the context of autonomous driving, much interest lies in methods that take advantage of the predictive response densities to produce safe and efficient end-to-end driving systems. End-to-end learners can identify multiple steering actions, but there is yet little research on how a potential steering action can be defined in terms of a probability distribution.
Predictive densities can also be used for in-depth analysis of crashes to further improve safety. When training an end-to-end learner we can systematically inspect observations with broad densities and provide more training data for areas of high (epistemic) uncertainty.

\bibliography{bibfile.bib}    

%% file: bibfile.bib
@Article{Skl1959,
  author  = {Sklar, A.},
  title   = {Fonctions de r{\'e}partition {\`a} $n$ dimensions et leurs marge},
  journal = {Publications de l'Institut de Statistique de l'Universit{\'e} de Paris},
  year    = {1959},
  volume  = {8},
  pages   = {229--231},
}

@InProceedings{RezMohWie2014, 
title = {Stochastic Backpropagation and Approximate Inference in Deep Generative Models}, 
author = {Danilo Jimenez Rezende and Shakir Mohamed and Daan Wierstra}, 
booktitle = {Proceedings of the 31st International Conference on Machine Learning}, 
pages = {1278--1286}, year = {2014}, editor = {Eric P. Xing and Tony Jebara}, volume = {32}
}

@inproceedings{KinWel2014,
  author    = {Diederik P. Kingma and
               Max Welling},
  editor    = {Yoshua Bengio and
               Yann LeCun},
  title     = {Auto-Encoding Variational {B}ayes},
  booktitle = {2nd International Conference on Learning Representations},
  year      = {2014}
}

@article{NotTanVilKoh2012,
author = { David J.   Nott  and  Siew   Li   Tan  and  Mattias   Villani  and  Robert   Kohn },
title = {Regression Density Estimation With Variational Methods and Stochastic Approximation},
journal = {Journal of Computational and Graphical Statistics},
volume = {21},
number = {3},
pages = {797--820},
year  = {2012}

}

@InProceedings{TitLaz2014, 
 title = {Doubly Stochastic Variational {B}ayes for non-Conjugate Inference}, 
 author = {Michalis Titsias and Miguel Lázaro-Gredilla}, 
 booktitle = {Proceedings of the 31st International Conference on Machine Learning}, 
 pages = {1971--1979}, 
 year = {2014}, 
 editor = {Eric P. Xing and Tony Jebara}, 
 volume = {32}
  }

@article{HofBleWanPai2013,
  author  = {Matthew D. Hoffman and David M. Blei and Chong Wang and John Paisley},
  title   = {Stochastic Variational Inference},
  journal = {Journal of Machine Learning Research},
  year    = {2013},
  volume  = {14},
  number  = {4},
  pages   = {1303--1347}
}

@article{OrmWan2010,
  title={Explaining variational approximations},
  author={Ormerod, John T and Wand, Matt P},
  journal={The American Statistician},
  volume={64},
  number={2},
  pages={140--153},
  year={2010},
  publisher={Taylor \& Francis}
}

@article{gneiting+r13,
author = "Gneiting, Tilmann and Ranjan, Roopesh",
journal = "Electronic Journal of Statistics",
pages = "1747--1782",
title = "Combining predictive distributions",
volume = "7",
year = "2013"
}

@article{gneiting+br07,
title = {Probabilistic forecasts, calibration and sharpness},
author = {Gneiting, Tilmann and Balabdaoui, Fadoua and Raftery, Adrian E.},
year = {2007},
journal = {Journal of the Royal Statistical Society Series B},
volume = {69},
number = {2},
pages = {243--268}
}

@Article{CarPol2010,
  author  = {Carvalho, Carlos M. and Polson, Nicholas, G.},
  title   = {The horseshoe estimator for sparse signals},
  journal = {Biometrica},
  year    = {2010},
  volume  = {97},
  pages   = {465--480},
}

@Article{CraSab2012,
  author  = {Craiu, V. R. and Sabeti, A.},
  title   = {In mixed company: {B}ayesian inference for bivariate conditional copula models with discrete and continuous outcomes},
  journal = {Journal of Multivariate Analysis},
  year    = {2012},
  volume  = {110},
  pages   = {106--120},
}

@article{Son2000,
	title={Multivariate dispersion models generated from {G}aussian copula},
	author={Song, Peter},
	journal={Scandinavian Journal of Statistics},
	volume={27},
	number={2},
	pages={305--320},
	year={2000},
	publisher={Wiley Online Library}
}

@article{RodPer2020,
  title={Beyond expectation: {D}eep joint mean and quantile regression for spatiotemporal problems},
  author={Filipe Rodrigues and F. Pereira},
  journal={IEEE Transactions on Neural Networks and Learning Systems},
  year={2020},
  volume={31},
  pages={5377--5389}
}

@inproceedings{UriMurLar2013,
 author = {Uria, Benigno and Murray, Iain and Larochelle, Hugo},
 booktitle = {Advances in Neural Information Processing Systems},
 editor = {C. J. C. Burges and L. Bottou and M. Welling and Z. Ghahramani and K. Q. Weinberger},
 pages = {2175–-2183},
 publisher = {Curran Associates, Inc.},
 title = {{RNADE}: {T}he real-valued neural autoregressive density-estimator},
 url = {https://proceedings.neurips.cc/paper/2013/file/53adaf494dc89ef7196d73636eb2451b-Paper.pdf},
 volume = {26},
 year = {2013}
}

@article{TraNguNotKoh2020,
author = {M.-N. Tran and N. Nguyen and D. Nott and R. Kohn},
title = {Bayesian deep net {GLM} and {GLMM}},
journal = {Journal of Computational and Graphical Statistics},
volume = {29},
number = {1},
pages = {97--113},
year  = {2020}
}

@inproceedings{KenGal2017,
 author = {Kendall, Alex and Gal, Yarin},
 booktitle = {Advances in Neural Information Processing Systems},
 editor = {I. Guyon and U. V. Luxburg and S. Bengio and H. Wallach and R. Fergus and S. Vishwanathan and R. Garnett},
 pages = {5574--5584},
 publisher = {Curran Associates, Inc.},
 title = {What uncertainties do we need in {B}ayesian deep learning for computer vision?},
 volume = {30},
 year = {2017}
}

@article{KleNotSmi2020,
    author={Nadja Klein and David J. Nott and Michael Stanley Smith},
    title={Marginally calibrated deep distributional regression},
    journal={Journal of Computational and Graphical Statistics},
    volume={30},
    number = {2},
    pages = {467--483},
    year ={2021}
}

@article{KleSmi2019,
 author = {Klein, Nadja and Smith, Michael Stanley},
 year = {2019},
 title = {Implicit copulas from {B}ayesian regularized regression smoothers},
 pages = {1143--1171},
 volume = {14},
 number = {4},
 issn = {1936-0975},
 journal = {Bayesian Analysis},
 doi = {10.1214/18-BA1138}
}

@article{SmiKle2020,
author = {Michael Stanley Smith and Nadja Klein},
title = {Bayesian inference for regression copulas},
journal = {Journal of Business \& Economic Statistics},
volume = {39},
number = {3},
pages = {712--728},
year  = {2021},
publisher = {Taylor & Francis}
}

@book{Nel2006,
 author = {Nelsen, R.},
 year = {2006},
 title = {An Introduction to Copulas},
 edition = {2nd},
 publisher = {Springer}
}

@article{BojTesDwo2016,
      title={End to end learning for self-driving cars}, 
      author={Mariusz Bojarski and Davide Del Testa and Daniel Dworakowski and Bernhard Firner and Beat Flepp and Prasoon Goyal and Lawrence D. Jackel and Mathew Monfort and Urs Muller and Jiakai Zhang and Xin Zhang and Jake Zhao and Karol Zieba},
      year={2016},
      note={arXiv:1604.07316},
      archivePrefix={arXiv},
      primaryClass={},
      pages = {1--4,6}
}

@inproceedings{MilFotDamAda2017,
 author = {Miller, Andrew and Foti, Nick and D\textquotesingle Amour, Alexander and Adams, Ryan P},
 booktitle = {Advances in Neural Information Processing Systems},
 editor = {I. Guyon and U. V. Luxburg and S. Bengio and H. Wallach and R. Fergus and S. Vishwanathan and R. Garnett},
 pages = {3--10},
 publisher = {Curran Associates, Inc.},
 title = {Reducing reparameterization gradient variance},
 volume = {30},
 year = {2017}
}

@article{CheSefKorXia2015,
  author={C. {Chen} and A. {Seff} and A. {Kornhauser} and J. {Xiao}},
  booktitle={2015 IEEE International Conference on Computer Vision (ICCV)}, 
  title={{D}eep{D}riving: {L}earning affordance for direct perception in autonomous driving}, 
  year={2015},
  volume={},
  number={},
  pages={2722--2730}}

@inproceedings{LakPriBlu2017,
 author = {Lakshminarayanan, Balaji and Pritzel, Alexander and Blundell, Charles},
 booktitle = {Advances in Neural Information Processing Systems},
 editor = {I. Guyon and U. V. Luxburg and S. Bengio and H. Wallach and R. Fergus and S. Vishwanathan and R. Garnett},
 pages = {},
 publisher = {Curran Associates, Inc.},
 title = {Simple and scalable predictive uncertainty estimation using deep ensembles},
 url = {https://proceedings.neurips.cc/paper/2017/file/9ef2ed4b7fd2c810847ffa5fa85bce38-Paper.pdf},
 volume = {30},
 year = {2017}
}

@InProceedings{BluCorKavWie2015,
  title = 	 {Weight uncertainty in neural networks},
  author = 	 {Blundell, Charles and Cornebise, Julien and Kavukcuoglu, Koray and Wierstra, Daan},
  booktitle = 	 {Proceedings of the 32nd International Conference on Machine Learning},
  pages = 	 {1613--1622},
  year = 	 {2015},
  editor = 	 {Bach, Francis and Blei, David},
  volume = 	 {37},
  series = 	 {Proceedings of Machine Learning Research},
  address = 	 {Lille, France},
  month = 	 {07--09 Jul},
  publisher =    {PMLR},
  url = 	 {
http://proceedings.mlr.press/v37/blundell15.html
}
}

@InProceedings{XuGaoYuTre2016,
author = {Xu, Huazhe and Gao, Yang and Yu, Fisher and Darrell, Trevor},
title = {End-to-end learning of driving models from large-scale video datasets},
booktitle = {Proceedings of the IEEE Conference on Computer Vision and Pattern Recognition (CVPR)},
pages = {1,7--8},
month = {July},
year = {2017}
}

@inproceedings{ChiMu2017,
author = {Chi, Lu and Mu, Yadong},
title = {Learning End-to-End Autonomous Steering Model from Spatial and Temporal Visual Cues},
year = {2017},
isbn = {9781450355063},
publisher = {Association for Computing Machinery},
address = {New York, NY, USA},
url = {https://doi.org/10.1145/3132734.3132737},
doi = {10.1145/3132734.3132737},
booktitle = {Proceedings of the Workshop on Visual Analysis in Smart and Connected Communities},
pages = {9--16},
numpages = {8},
location = {Mountain View, California, USA},
series = {VSCC '17}
}

@article{AmiSolKarRus2018,
    title={Spatial uncertainty sampling for end-to-end control},
    author={Alexander Amini and Ava Soleimany and Sertac Karaman and Daniela Rus},
    year={2019},
    eprint={1805.04829},
    note = {arXiv:1805.04829},
    archivePrefix={arXiv},
    primaryClass={},
    pages = {1--5}
}

@InProceedings{GalGha2016,
  title = 	 {Dropout as a {B}ayesian approximation: representing model uncertainty in deep learning},
  author = 	 {Gal, Yarin and Ghahramani, Zoubin},
  booktitle = 	 {Proceedings of The 33rd International Conference on Machine Learning},
  pages = 	 {1050--1059},
  year = 	 {2016},
  editor = 	 {Balcan, Maria Florina and Weinberger, Kilian Q.},
  volume = 	 {48},
  series = 	 {Proceedings of Machine Learning Research},
  address = 	 {New York, New York, USA},
  month = 	 {20--22 Jun},
  publisher =    {PMLR},
  url = 	 {http://proceedings.mlr.press/v48/gal16.html}
}

@article{MicKwiGal2018,
    title={Evaluating uncertainty quantification in end-to-end autonomous driving control},
    author={Rhiannon Michelmore and Marta Kwiatkowska and Yarin Gal},
    year={2018},
    eprint={1811.06817},
    note = {arXiv:1811.06817},
    archivePrefix={arXiv},
    primaryClass={},
    pages = {1--5}
}

@InProceedings{SnoRipSwe2015,
  title = 	 {Scalable Bayesian Optimization Using Deep Neural Networks},
  author = 	 {Snoek, Jasper and Rippel, Oren and Swersky, Kevin and Kiros, Ryan and Satish, Nadathur and Sundaram, Narayanan and Patwary, Mostofa and Prabhat, Mr and Adams, Ryan},
  booktitle = 	 {Proceedings of the 32nd International Conference on Machine Learning},
  pages = 	 {2171--2176},
  year = 	 {2015},
  editor = 	 {Bach, Francis and Blei, David},
  volume = 	 {37},
  series = 	 {Proceedings of Machine Learning Research},
  address = 	 {Lille, France},
  month = 	 {07--09 Jul},
  publisher =    {PMLR},
  url = 	 {
http://proceedings.mlr.press/v37/snoek15.html
}
}

@article{RiqTucSno2018,
      title={Deep {B}ayesian bandits showdown: {A}n empirical comparison of {B}ayesian deep networks for {T}hompson sampling}, 
      author={Carlos Riquelme and George Tucker and Jasper Snoek},
      year={2018},
      eprint={1802.09127},
      note = {arXiv:1802.09127},
      archivePrefix={arXiv},
      primaryClass={},
      pages = {4}
}

@inproceedings{PinGorNal2019,
 author = {Pinsler, Robert and Gordon, Jonathan and Nalisnick, Eric and Hern\'{a}ndez-Lobato, Jos\'{e} Miguel},
 booktitle = {Advances in Neural Information Processing Systems},
 editor = {H. Wallach and H. Larochelle and A. Beygelzimer and F. d\textquotesingle Alch\'{e}-Buc and E. Fox and R. Garnett},
 pages = {7--8},
 publisher = {Curran Associates, Inc.},
 title = {Bayesian batch active learning as sparse subset approximation},
 url = {https://proceedings.neurips.cc/paper/2019/file/84c2d4860a0fc27bcf854c444fb8b400-Paper.pdf},
 volume = {32},
 year = {2019}
}

@article{ObeRas2019,
      title={Benchmarking the neural linear model for regression}, 
      author={Sebastian W. Ober and Carl Edward Rasmussen},
      year={2019},
      eprint={1912.08416},
      note = {arXiv:1912.08416},
      archivePrefix={arXiv},
      primaryClass={},
      pages = {5}
}

@inproceedings{WilGha2010,
 author = {Wilson, Andrew G and Ghahramani, Zoubin},
 booktitle = {Advances in Neural Information Processing Systems},
 editor = {J. Lafferty and C. Williams and J. Shawe-Taylor and R. Zemel and A. Culotta},
 pages = {},
 publisher = {Curran Associates, Inc.},
 title = {Copula processes},
 url = {https://proceedings.neurips.cc/paper/2010/file/fc8001f834f6a5f0561080d134d53d29-Paper.pdf},
 volume = {23},
 year = {2010}
}

@article{KleKne2016b,
 author = {Klein, N. and Kneib, T.},
 year = {2016},
 title = {Simultaneous inference in structured additive conditional copula regression models: A unifying {B}ayesian approach},
 pages = {841--860},
 volume = {26},
 number = {4},
 journal = {Statistics and Computing},
 doi = {10.1007/s11222-015-9573-6}
}

@article{PitChaKoh2006,
 author = {Pitt, M. and Chan, D. and Kohn, R.},
 year = {2006},
 title = {Efficient {B}ayesian inference for {G}aussian copula regression models},
 pages = {537--554},
 volume = {93},
 issn = {0006-3444},
 journal = {Biometrika}
}

@article{SonLiYua2009,
 author = {Song, Peter X-K and Li, Mingyao and Yuan, Ying},
 year = {2009},
 title = {Joint regression analysis of correlated data using {G}aussian copulas},
 url = {https://onlinelibrary.wiley.com/doi/10.1111/j.1541-0420.2008.01058.x},
 pages = {60--68},
 volume = {65},
 number = {1},
 issn = {0006-341X},
 journal = {Biometrics},
 doi = {10.1111/j.1541-0420.2008.01058.x}
}

@InProceedings{PeaZakBri2018,
  title = 	 {High-Quality Prediction Intervals for Deep Learning: A Distribution-Free, Ensembled Approach},
  author =       {Pearce, Tim and Brintrup, Alexandra and Zaki, Mohamed and Neely, Andy},
  booktitle = 	 {Proceedings of the 35th International Conference on Machine Learning},
  pages = 	 {4088},
  year = 	 {2018},
  editor = 	 {Dy, Jennifer and Krause, Andreas},
  volume = 	 {80},
  series = 	 {Proceedings of Machine Learning Research},
  month = 	 {10--15 Jul},
  publisher =    {PMLR},
  url = 	 {http://proceedings.mlr.press/v80/pearce18a.html}
}

@INPROCEEDINGS{AbbHajKar2019,
  author={Abbasi, Sajjad and Hajabdollahi, Mohsen and Karimi, Nader and Samavi, Shadrokh},
  booktitle={2020 International Conference on Machine Vision and Image Processing (MVIP)}, 
  title={Modeling teacher-student techniques in deep neural networks for knowledge distillation}, 
  year={2020},
  volume={},
  number={},
  pages={1--6},
  doi={10.1109/MVIP49855.2020.9116923}}

@article{GneBalRaf2007,
author = {Gneiting, Tilmann and Balabdaoui, Fadoua and Raftery, Adrian},
year = {2007},
title = {Probabilistic forecasts, calibration and sharpness},
journal = {Journal of the Royal Statistical Society: Series B (Statistical Methodology)},
volume = {69},
number = {2},
pages = {243--268},
}

@InProceedings{GupPleSun2017, 
title = {On Calibration of Modern Neural Networks}, 
author = {Chuan Guo and Geoff Pleiss and Yu Sun and Kilian Q. Weinberger}, 
booktitle = {Proceedings of the 34th International Conference on Machine Learning}, 
pages = {1321--1330}, 
year = {2017}, 
editor = {Precup, Doina and Teh, Yee Whye}, 
volume = {70}, 
series = {Proceedings of Machine Learning Research}, 
month = {06--11 Aug}, 
publisher = {PMLR}, 
url = { http://proceedings.mlr.press/v70/guo17a.html }}

@inproceedings{HeLakYee2020,
 author = {He, Bobby and Lakshminarayanan, Balaji and Teh, Yee Whye},
 booktitle = {Advances in Neural Information Processing Systems},
 editor = {H. Larochelle and M. Ranzato and R. Hadsell and M. F. Balcan and H. Lin},
 pages = {1010--1022},
 publisher = {Curran Associates, Inc.},
 title = {Bayesian Deep Ensembles via the Neural Tangent Kernel},
 url = {https://proceedings.neurips.cc/paper/2020/file/0b1ec366924b26fc98fa7b71a9c249cf-Paper.pdf},
 volume = {33},
 year = {2020}
}

@inproceedings{LinZhaHsu2018,
author = {Lin, Shih-Chieh and Zhang, Yunqi and Hsu, Chang-Hong and Skach, Matt and Haque, Md E. and Tang, Lingjia and Mars, Jason},
title = {The architectural implications of autonomous driving: constraints and acceleration},
year = {2018},
isbn = {9781450349116},
publisher = {Association for Computing Machinery},
address = {New York, NY, USA},
booktitle = {Proceedings of the Twenty-Third International Conference on Architectural Support for Programming Languages and Operating Systems},
pages = {751--766},
numpages = {16},
location = {Williamsburg, VA, USA},
series = {ASPLOS '18}
}

@incollection{Nea2011,
   title={{MCMC} using {H}amiltonian dynamics},
   booktitle = {Handbook of Markov Chain Monte Carlo},
   editor = {Brooks, Steve and Gelman, Andrew and Jones, Galin and Meng, Xiao-Li},
   author={Radford M. Neal},
   ISBN={9780429138508},
   url={http://dx.doi.org/10.1201/b10905},
   DOI={10.1201/b10905},
   publisher={Chapman and Hall/CRC},
   pages = {1--8, 29--30},
   year={2011},
   month={May}
}

@article{OngNotSmi2017,
author = {Victor M.-H. Ong and David J. Nott and Michael S. Smith},
title = {Gaussian variational approximation with a factor covariance structure},
journal = {Journal of Computational and Graphical Statistics},
volume = {27},
number = {3},
pages = {465, 470--474, 476-478},
year  = {2018},
publisher = {Taylor & Francis},
doi = {10.1080/10618600.2017.1390472},
URL = { 
        https://doi.org/10.1080/10618600.2017.1390472},
eprint = { https://doi.org/10.1080/10618600.2017.1390472}
}

@article{Zei2012,
      title={{ADADELTA}: an adaptive learning rate method}, 
      author={Matthew D. Zeiler},
      year={2012},
      eprint={1212.5701},
      note = {arXiv:1212.5701},
      archivePrefix={arXiv},
      primaryClass={},
      pages = {3--4}
}

@article{PolSok2017,
   title={Deep learning: a {B}ayesian perspective},
   volume={12},
   number={4},
   journal={Bayesian Analysis},
   publisher={Institute of Mathematical Statistics},
   author={Polson, Nicholas G. and Sokolov, Vadim},
   year={2017},
   month={12},
   pages={1275--1304}
}

@article{KleKne2016a,
author = {Nadja Klein and Thomas Kneib},
title = {{Scale-dependent priors for variance parameters in structured additive distributional regression}},
volume = {11},
journal = {Bayesian Analysis},
number = {4},
publisher = {International Society for Bayesian Analysis},
pages = {1071 -- 1106},
year = {2016},
doi = {10.1214/15-BA983},
URL = {https://doi.org/10.1214/15-BA983}
}

@article{SchEdeHad2018,
      title={A commute in data: the comma2k19 dataset}, 
      author={Harald Schafer and Eder Santana and Andrew Haden and Riccardo Biasini},
      year={2018},
      eprint={1812.05752},
      note = {arXiv:1812.05752},
      archivePrefix={arXiv},
      primaryClass={},
      pages = {1}
}

@unpublished{Bis1994,
            type = {Technical Report},
           title = {Mixture density networks},
          author = {Christopher M. Bishop},
         address = {Birmingham},
       publisher = {Aston University},
            year = {1994},
             url = {https://publications.aston.ac.uk/id/eprint/373/},
             note = {unpublished}
}

@book{GooBenCou2016,
    title={Deep Learning},
    author={Ian Goodfellow and Yoshua Bengio and Aaron Courville},
    publisher={MIT Press},
    note={http://www.deeplearningbook.org},
    year={2016}
}

@InProceedings{KulFenErm2018,
  title = 	 {Accurate uncertainties for deep learning using calibrated regression},
  author =       {Kuleshov, Volodymyr and Fenner, Nathan and Ermon, Stefano},
  booktitle = 	 {Proceedings of the 35th International Conference on Machine Learning},
  pages = 	 {2796--2804},
  year = 	 {2018},
  editor = 	 {Dy, Jennifer and Krause, Andreas},
  volume = 	 {80},
  series = 	 {Proceedings of Machine Learning Research},
  month = 	 {10--15 Jul},
  publisher =    {PMLR},
  url = 	 {http://proceedings.mlr.press/v80/kuleshov18a.html}
}

@misc{ZhaDalSab2019,
      title={Confidence Calibration for Convolutional Neural Networks Using Structured Dropout}, 
      author={Zhilu Zhang and Adrian V. Dalca and Mert R. Sabuncu},
      year={2019},
      eprint={1906.09551},
      archivePrefix={arXiv},
      note = {arXiv:1906.09551},
      primaryClass={cs.LG}
}

@article{RumHinWil1986,
  title={Learning representations by back-propagating errors},
  author={D. Rumelhart and Geoffrey E. Hinton and Ronald J. Williams},
  journal={Nature},
  year={1986},
  volume={323},
  pages={533-536}
}

@article{Rac2008,
url = {http://dx.doi.org/10.1561/0800000009},
year = {2008},
volume = {3},
journal = {Foundations and Trends® in Econometrics},
title = {Nonparametric Econometrics: A Primer},
doi = {10.1561/0800000009},
issn = {1551-3076},
number = {1},
pages = {1-88},
author = {Jeffrey S. Racine}
}

@article{PiiVeh2017,
   title={Sparsity information and regularization in the horseshoe and other shrinkage priors},
   volume={11},
   number={2},
   journal={Electronic Journal of Statistics},
   publisher={Institute of Mathematical Statistics},
   author={Piironen, Juho and Vehtari, Aki},
   year={2017},
   pages={5024}
}

@misc{MarDuh2020,
      title={keras-mdn-layer}, 
      author={Charles Martin and Douglas Duhaime },
      year={2020},
      url = {https://github.com/cpmpercussion/keras-mdn-layer},
      note = {published with MIT license}
}

@misc{Cho2015,
  title={Keras},
  author={Chollet, Fran\c{c}ois and others},
  year={2015},
  howpublished={\url{https://keras.io}},
}

@misc{Aba2015,
title={ {TensorFlow}: Large-Scale Machine Learning on Heterogeneous Systems},
url={https://www.tensorflow.org/},
note={Software available from tensorflow.org},
author={
    Mart\'{\i}n~Abadi and
    Ashish~Agarwal and
    Paul~Barham and
    Eugene~Brevdo and
    Zhifeng~Chen and
    Craig~Citro and
    Greg~S.~Corrado and
    Andy~Davis and
    Jeffrey~Dean and
    Matthieu~Devin and
    Sanjay~Ghemawat and
    Ian~Goodfellow and
    Andrew~Harp and
    Geoffrey~Irving and
    Michael~Isard and
    Yangqing Jia and
    Rafal~Jozefowicz and
    Lukasz~Kaiser and
    Manjunath~Kudlur and
    Josh~Levenberg and
    Dandelion~Man\'{e} and
    Rajat~Monga and
    Sherry~Moore and
    Derek~Murray and
    Chris~Olah and
    Mike~Schuster and
    Jonathon~Shlens and
    Benoit~Steiner and
    Ilya~Sutskever and
    Kunal~Talwar and
    Paul~Tucker and
    Vincent~Vanhoucke and
    Vijay~Vasudevan and
    Fernanda~Vi\'{e}gas and
    Oriol~Vinyals and
    Pete~Warden and
    Martin~Wattenberg and
    Martin~Wicke and
    Yuan~Yu and
    Xiaoqiang~Zheng},
  year={2015},
}
